\def\year{2021}\relax
\documentclass[letterpaper]{article} 
\usepackage{aaai21}  
\usepackage{times}  
\usepackage{helvet} 
\usepackage{courier}  
\usepackage[hyphens]{url}  
\usepackage{graphicx} 
\usepackage{algorithm}
\usepackage[noend]{algpseudocode}
\usepackage{multicol}
\usepackage{natbib}  
\usepackage{caption} 
\usepackage{amsmath}               
  {
      
      \newtheorem{strategy}{Strategy}
      \newtheorem{definition}{Definition}
      
  }
\usepackage{subcaption}
\usepackage{tcolorbox}
\usepackage{color}
\definecolor{myred}{rgb}{0.8,0,0}
\definecolor{mygreen}{rgb}{0,0.6,0}
\definecolor{myblue}{rgb}{0,0,0.7}
\definecolor{myyellow}{rgb}{0,0.6,0.7}

\title{Selection-Expansion:\\
A Unifying Framework for Motion-Planning and Diversity Search Algorithms }
\author{
    Alexandre Chenu\textsuperscript{\rm 1}, \\
    Nicolas Perrin-Gilbert\textsuperscript{\rm 1}, 
    Stéphane Doncieux\textsuperscript{\rm 1},
    Olivier Sigaud\textsuperscript{\rm 1}
    \\
}
\affiliations{
    \textsuperscript{\rm 1}Sorbonne Université, CNRS, Institut des Systèmes Intelligents et de Robotique, ISIR F-75005 Paris, France\\
    chenu@isir.upmc.fr

}

\begin{document}
\maketitle
\begin{abstract}
Reinforcement learning agents need a reward signal to learn successful policies. When this signal is sparse or the corresponding gradient is deceptive, such agents need a dedicated mechanism to efficiently explore their search space without relying on the reward. Looking for a large diversity of behaviors or using Motion Planning (MP) algorithms are two options in this context. In this paper, we build on the common roots between these two options to investigate the properties of two diversity search algorithms, the Novelty Search and the Goal Exploration Process algorithms. These algorithms look for diversity in an {\em outcome space} or {\em behavioral space} which is generally hand-designed to represent what matters for a given task. The relation to MP algorithms reveals that the smoothness, or lack of smoothness of the mapping between the policy parameter space and the outcome space plays a key role in the search efficiency.
In particular, we show empirically that, if the mapping is smooth enough, i.e. if two close policies in the parameter space lead to similar outcomes, then diversity algorithms tend to inherit exploration properties of MP algorithms. By contrast, if it is not, diversity algorithms lose these properties and their performance strongly depends on specific heuristics, notably filtering mechanisms that discard some of the explored policies.
\end{abstract}

\section{Introduction}
Deep Reinforcement learning (RL) and Deep Neuro-Evolution (NE) methods have recently undergone outstanding progress, obtaining more and more impressive performance in games and robotics applications \cite{DRL_rob,DRL_mujoco,deep_neuroevo,silver2016mastering,silver2017mastering,andrychowicz2018learning,vinyals2019grandmaster,akkaya2019solving}.
However, despite these successes, some fundamental difficulties remain in {\em hard exploration problems}. First, when the reward signal is sparse or when the corresponding gradient is deceptive, RL agents cannot rely on the reward signal to steer their learning process, resulting in complete failure or poor performance \citep{matheron2019problem}. Besides, RL agents struggle when only complex trajectories can reach the target region, as for example in a complicated maze, and when such target-reaching agents correspond to a very small domain of the policy parameter space \citep{ecoffet2019goexplore,matheron2020pbcs}.
In such contexts, combining Deep RL with algorithms explicitly designed to look for diversity in a relevant search space has several attractive properties. More precisely, it has been hypothesized in \citep{doncieux_novelty_2019} that defining an {\em outcome space}\footnote{Also called {\em behavioral space} in the literature} as the space that matters to determine whether a policy is successful and looking for diversity in that space might be the best option to tackle the sparse reward exploration problem. 

We can distinguish two classes of algorithms to implement this diversity search approach: Goal Exploration Process (GEP) \citep{forestier_modular_2016,forestier_these_2019,benureau_behavioral_2016} and Novelty Search (NS) \citep{lehman_abandoning_2011}. The former has been combined with RL in \cite{colas2016} whereas the latter is used in the same way in \cite{cideron2020qdrl}.

This paper investigates the properties of these two classes of algorithms. In a first part, based on a very general \textit{selection-expansion} framework, we reveal a similarity between these algorithms and Motion Planning (MP) algorithms like Expansive Spaces Trees (EST) \cite{LatombePath} and Rapidly-exploring Random Trees (RRT) \cite{Lavalle98rapidly-exploringrandom}. In a second part, we empirically compare both algorithms in two environments where a 
smoothness assumption on which MP algorithms implicitly rely either holds or not. We show that diversity algorithms are highly dependent on the design of the outcome space where the search for diversity is performed, and that the smoothness of the mapping between the policy parameter space and the outcome space plays a key role in their search dynamics. In particular, we show that if the mapping is smooth enough, GEP and NS inherit the exploration properties of their MP counterparts and GEP outperforms NS. By contrast, if it is not, which is the usual case, NS and GEP perform differently and their performance strongly depends on specific heuristics, notably filtering mechanisms that discard some of the explored policies. 

\section{Methods}
In this section we highlight that NS and GEP share properties with two well-known MP algorithms, EST and RRT. To establish the similarity between both families of algorithms, we start from a more general framework that we call {\em selection-expansion} algorithms.

\subsection{Selection-expansion algorithms}
\label{sec:sel_exp}

Imagine an agent searching in some space and looking for an area it knows nothing about. What should it do? 
The most classical approach is to keep a memory of what has already been explored, and to progress locally, i.e. by reconsidering previous trajectories or behaviors, and by expanding or slightly modifying them to find new areas of the space to explore. This is the basis of virtually all sampling-based motion planning algorithms, and the core mechanism of Go-Explore \citep{ecoffet2019goexplore}.
We call this kind of algorithms selection-expansion algorithms because they share the common structure of maintaining an archive of previous samples and iterating over a sequence of two operators:

\begin{itemize}
    \item 
    the {\bf selection operator} that chooses in the archive a sample from which to expand;
    \item
    the {\bf expansion operator} that adds one or several new samples built from the selected sample.
\end{itemize}
\noindent
Usually, selection and expansion operators are designed to efficiently expand the frontier of explored areas towards unexplored regions of the space.
To do so, there are two popular selection strategies.
One can either:
    \begin{strategy}
    \label{strat:est}
rank all elements in the archive in terms of distance to their neighbors, and preferentially select those far away from their neighbors, which suggests that they lie in a region with a low density of exploration; or
    \end{strategy}
        \begin{strategy}
    \label{strat:rrt}
     randomly draw a sample anywhere in the search space and select the closest sample in the archive. This way, samples which are close to large unexplored regions have a higher chance of being selected.
   \end{strategy}

In the next section, we describe applications of the above selection-expansion algorithms in two domains, namely Motion Planning and Diversity Search algorithms. This reveals a striking similarity between both families of algorithms.

\subsection{Application to Motion Planning}

\begin{figure}[h!]
     \centering
     \includegraphics[width = 0.9\hsize]{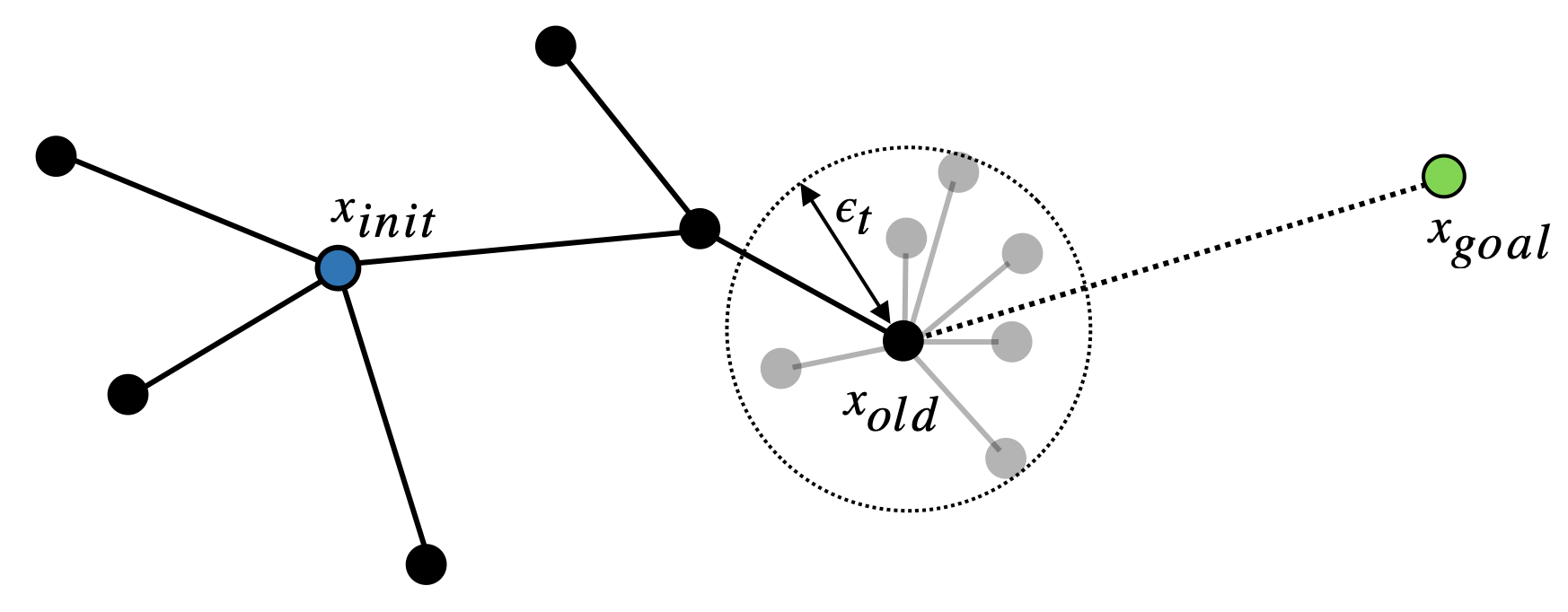}
    \label{fig:random_expansion}
    \caption{Expansion operators in motion planning with unknown dynamical systems. When the dynamical system is unknown, random controls are propagated through the system, yielding random nodes contained in a local ball.}
    \label{fig:expansion_MP}
\end{figure}

\begin{figure*}[ht!]
    \centering
    \begin{subfigure}[ht]{0.27\textwidth}
         \centering
         \includegraphics[width =\hsize]{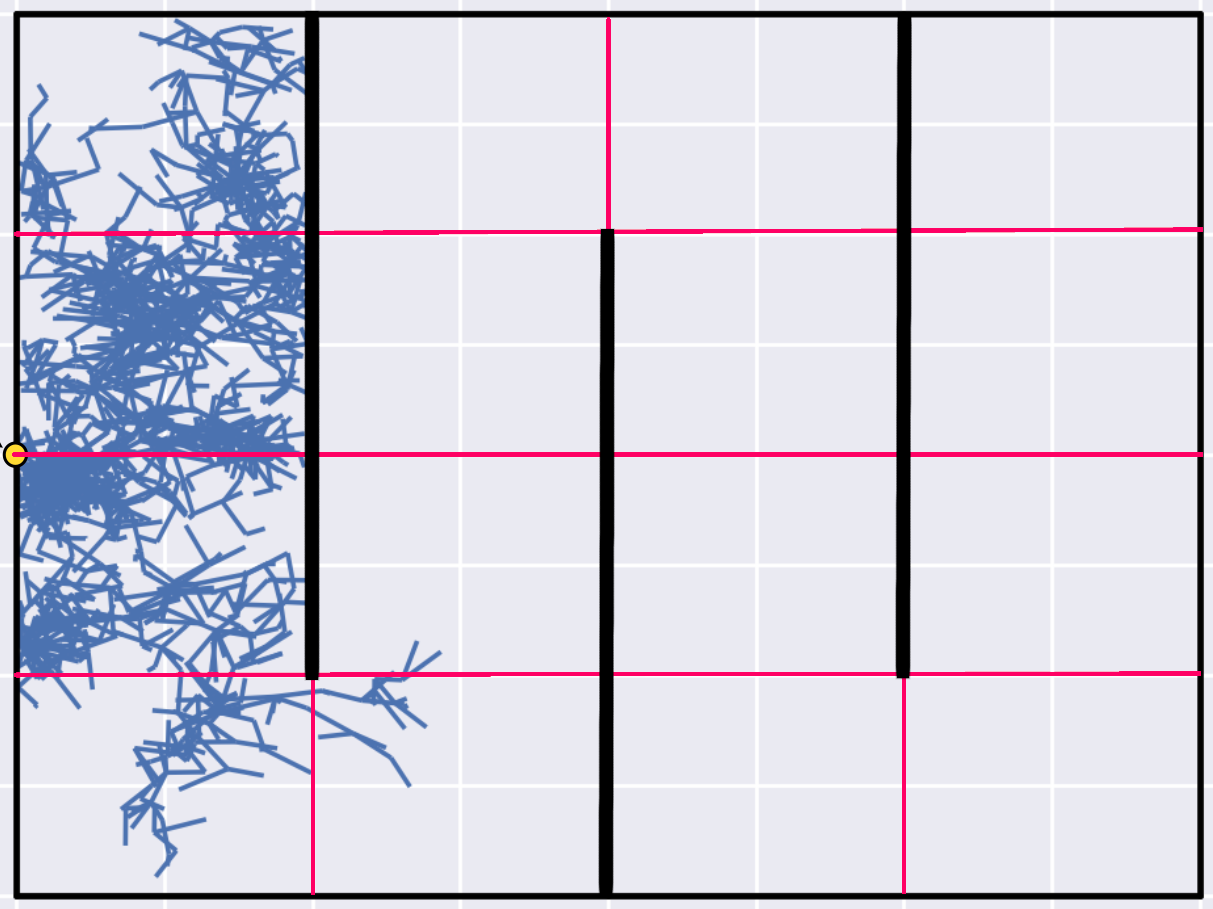}
         \caption{EST Search tree.}
        \label{fig:EST_marble}
    \end{subfigure}
    \hfill
    \begin{subfigure}[ht]{0.27\textwidth}
         \centering
         \includegraphics[width = \hsize]{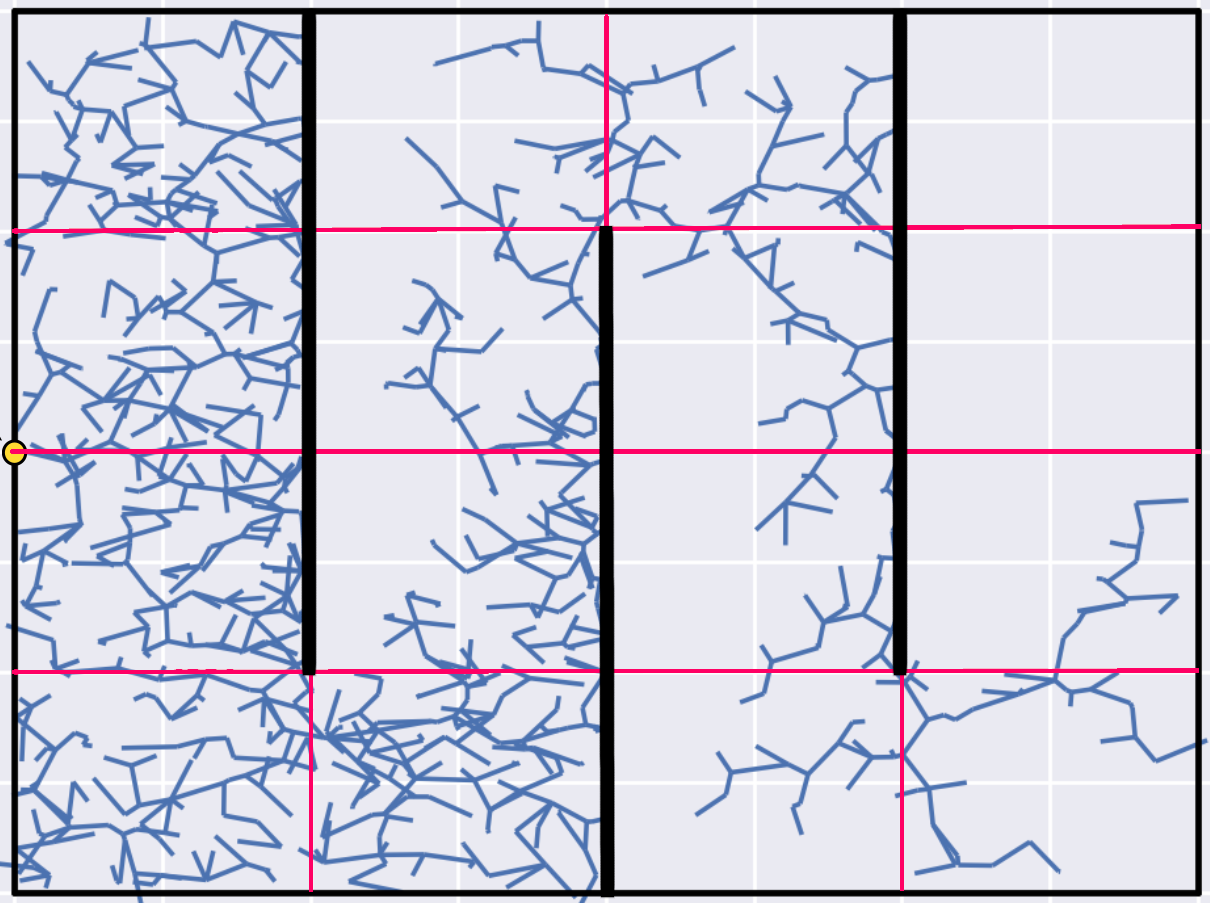}
         \caption{RRT Search tree.}
        \label{fig:RRT_marble}
    \end{subfigure}
    \hfill
    \begin{subfigure}[ht]{0.29\textwidth}
         \centering
         \includegraphics[width = \hsize]{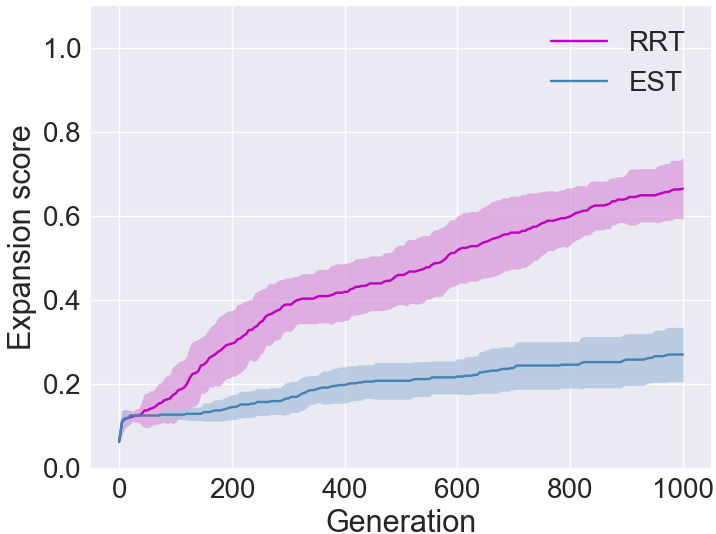}
         \caption{Expansion scores.}
        \label{fig:MP_exp_cov_marble}
    \end{subfigure}
        \caption{Empirical comparison of EST and RRT. The SimpleMaze environment is divided into a $4 \times 4$ grid to compute expansion scores of the MP algorithms. Search trees are shown after 1000 iterations. The means and standard deviations of the expansion scores are computed over 30 runs.}
        \label{fig:RRT_EST_marble}
\end{figure*}

In Motion Planning (MP), the goal is to find a trajectory for a system navigating from a starting configuration to a goal configuration or region. More formally, MP problems can be defined as follows:  

\begin{definition}
A motion-planning problem is defined by a $(C_{free},s_{0},C_{target})$ triplet, where $C_{free}$ is the space of free configurations, $s_{0}$ is a starting configuration and $C_{target}$ is the space of target configurations. The goal is to find a valid trajectory $\tau = (s_{i})_{i=0,\dots,n} $ between $s_{0}$ and $C_{target}$.
\end{definition}

Some MP algorithms use selection-expansion algorithms to build an exploring tree eventually containing a path from the starting configuration to the goal. 
Nodes of the graph are configurations, and edges represent the fact that the system can navigate between two nodes. Thus, in the MP context, the need for a local expansion operator comes from the fact that the system must navigate locally from its current configuration to the next.

When the model of the system is known, finding controls to navigate between two nodes can be easy, and two successive nodes can potentially be far away from each other. We do not consider this case here. Instead, we focus on the case where the model of the system is unknown, and assume that the effects of the dynamics are not easily predictable or controllable.
In that case, one must call upon random actions for a few time steps and rely on the fact that, if many random actions are tried, interesting motions may occur, as shown in \figurename~\ref{fig:expansion_MP}.
Thus, the expansion operator of such ``model-free" MP algorithms typically performs a random action from the selected configuration to reach a new configuration then added to the exploration tree. 

For the selection operator, there exist MP algorithms corresponding to both strategies described above.  

\subsubsection{Expansive Spaces Trees}

Expansive Spaces Trees (EST) corresponds to a family of algorithms where the selection operator uses Strategy~\ref{strat:est}. These algorithms select the most isolated nodes based on an estimate of the local density of nodes. Various approximations of the local density can be used.
For instance, a node can be selected based on its number of neighbors within a certain range $D$. 
The nodes are selected with a probability distribution based on the weight of nodes so that the nodes with fewer neighbors tend to be selected with higher frequency than others. 

Other estimates of the local density of nodes can also be used. In this paper, we consider the mean distance to the K-nearest nodes as an estimation of the local density. Given a set of N nodes $S = \{s_{i}\}_{[1,N]} \in C_{free}^{N}$ and a set of K nearest-neighbors $\{\mu_{1},...,\mu_{k}\}\subset S$ associated to node $s_{n}$, the latter has a weight :

\begin{equation}
\label{eq:est_weight}
    w_{n} = \frac{1}{k}\sum_{i=1}^{k}{dist(s_{n},\mu_{i}}).
\end{equation}
\noindent

The probability $p_{n}$ for $s_{n}$ to be selected is proportional to its weight:

\begin{equation}
\label{eq:est_selec}
    p_n = \frac{w_{n}}{\sum_{i=0}^{N}w_{i}}.
\end{equation}

Besides, in the general case without specific knowledge on the system, a random control input is used during one or few steps to expand $s_n$ to a new state $s_{new}$. If no collision occurs, $s_{new}$ is added to the tree. 

\subsubsection{Rapidly-exploring Random Trees}
\label{sec:rrt}

Like EST, Rapidly-exploring Random Trees (RRT) is a sampling-based path-planning algorithm. But, in contrast to EST, RRT performs selection according to Strategy~\ref{strat:rrt}. That is, it draws a random goal configuration $s_{samp}$ and selects the closest node in the set of already visited nodes. Note that sampling a random configuration requires to determine the boundaries of the space where to sample from, a stronger prerequisite than in EST.

 \begin{figure}[!ht]
    \centering
    \includegraphics[width = 0.6\hsize]{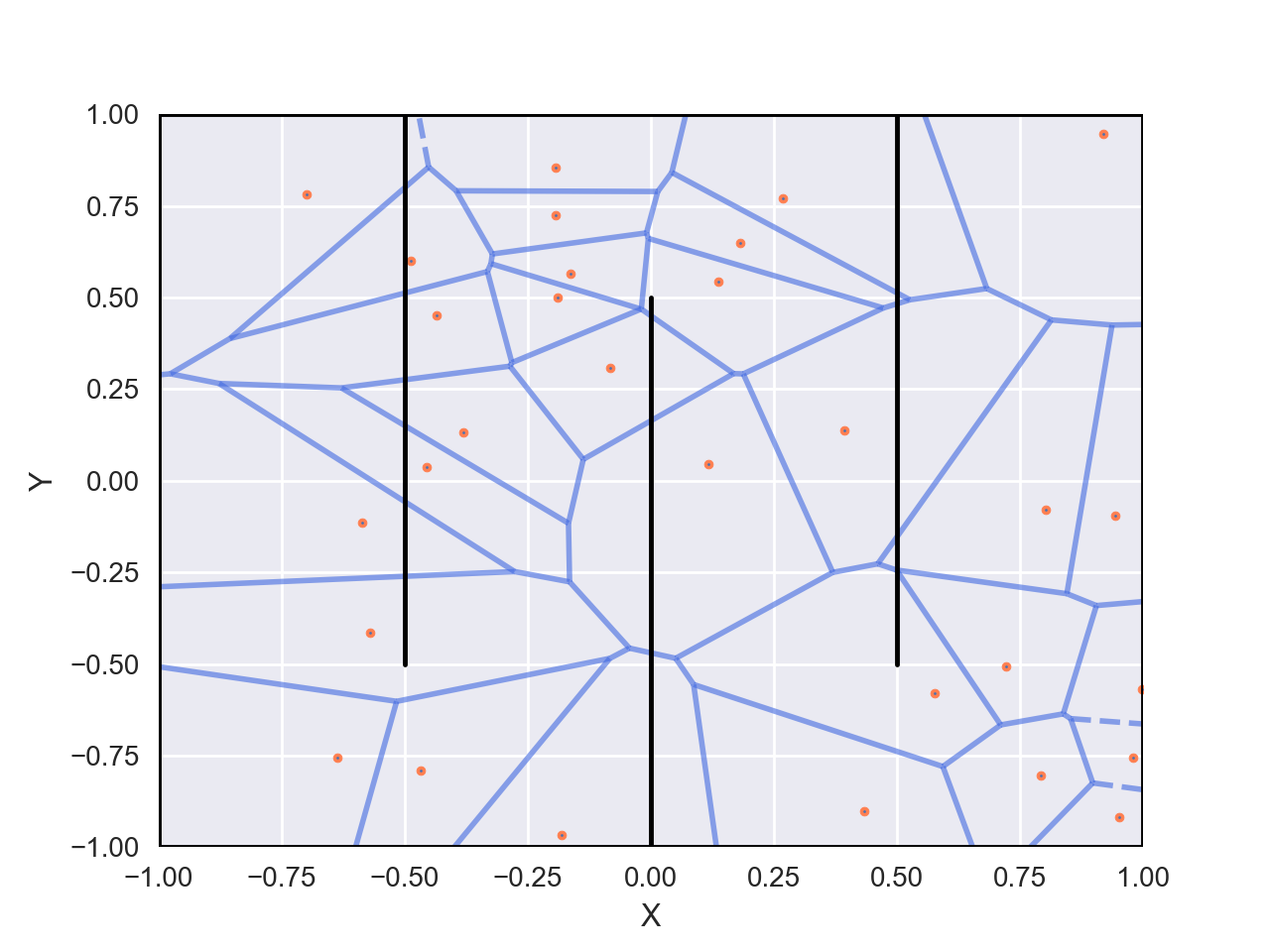}
    \caption{Voronoi diagram for 30 configurations in a simple 2D maze. With RRT, if these configurations are the current set of nodes, the probability to select a node is proportional to the volume of the corresponding Voronoi cell.}
    \label{fig:voronoi_diagram_simple}
\end{figure}

Given a set of nodes $\{s_{i}\}_{i\in[1,N]} \in C_{free}^{N}$, one can define the {\em Voronoi diagram} of these points as a set of {\em Voronoi cells} with one Voronoi cell per point, where the Voronoi cell of each point $s_i$ is the subspace of all points that are closer to $s_i$ than to any other point of the set. An example of Voronoi diagram is depicted in \figurename~\ref{fig:voronoi_diagram_simple}.
When selecting randomly, the probability $p_k$ for an already visited node $s_k$ to be selected is proportional to the volume of its Voronoi cell: 
\begin{equation}
\label{eq:rrt_selec}
    p_k = \frac{volume(Voronoi\ cell\ s_{k})}{\sum_{i=0}^{N}volume(Voronoi\ cell\ s_{i})}.
\end{equation}
After selection, without knowledge on the system, expansion is also performed by applying a random control. 


\subsubsection{Comparative search properties of EST and RRT}
\label{sec:compar_EST_RRT}

We empirically compare the exploration properties of the selection operators of EST and RRT in the ``SimpleMaze" environment which is further described in Section~\ref{sec:setup}. The pseudo-codes of both algorithms are given in Appendix~\ref{sec:algo_MP}.

To assess expansion, we divide the maze into a $4\times4$ expansion grid shown in \figurename~\ref{fig:RRT_EST_marble}. The expansion score is the number of zones containing at least one node over the total number of zones, i.e $16$. 

Both algorithms start with a single initial node in the middle of the left side (coordinates $(-1,0)$, see \figurename~\ref{fig:voronoi_diagram_simple}).
Figures~\ref{fig:EST_marble} and~\ref{fig:RRT_marble} display exploration trees for both EST and RRT after 1000 iterations.

The evolution of expansion presented in Figure~\ref{fig:MP_exp_cov_marble} shows that RRT explores the maze faster than EST. 


\begin{figure*}[ht!]
    \centering
    \begin{subfigure}[ht]{0.49\textwidth}
         \centering
         \includegraphics[width = \hsize]{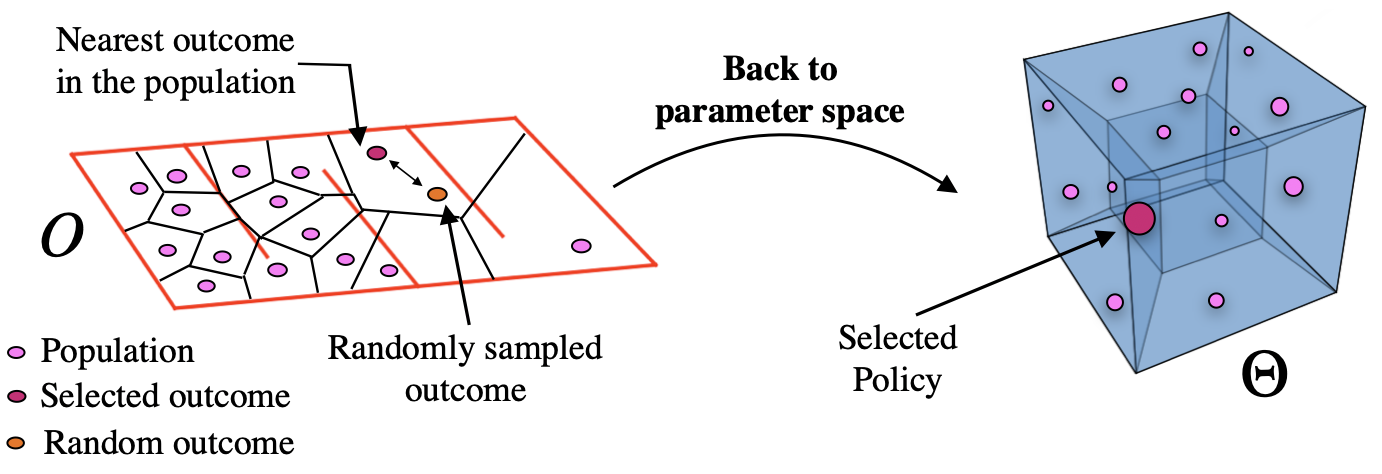}
         \caption{Goal-Exploration Process}
        \label{fig:selection_GEP}
    \end{subfigure}
    \hfill
    \begin{subfigure}[ht]{0.49\textwidth}
         \centering
         \includegraphics[width = \hsize]{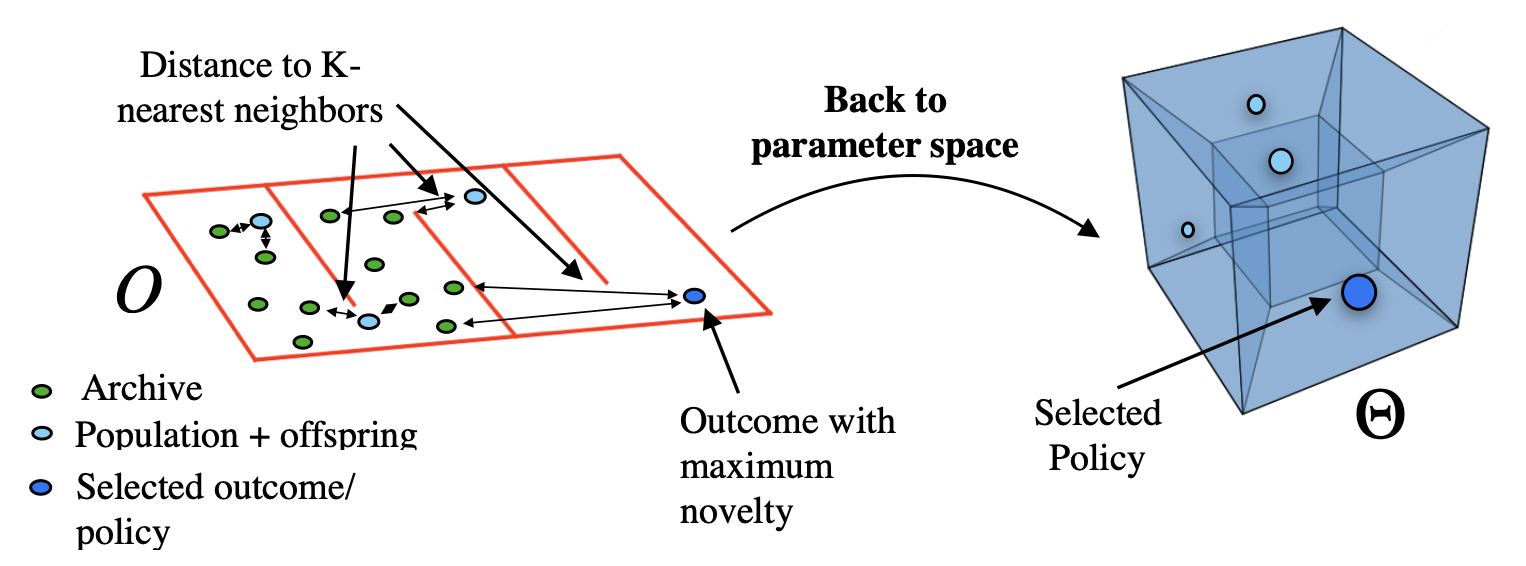}
         \caption{Novelty Search}
        \label{fig:selection_NS}
    \end{subfigure}
        \caption{Selection in GEP and NS. In GEP, an outcome is randomly sampled and the policy yielding the closest outcome is selected. In NS, the novelty is computed w.r.t to the archive. The policies yielding the most novel outcomes are selected.}
        \label{fig:selection_GEP_NS}
\end{figure*}

\subsection{Application to diversity search algorithms}
\label{sec:appli_div_search}

We now turn to the policy search context. In policy search, we consider a parametric policy $\pi_\theta$ where $\theta$ is a vector of parameters in a policy parameter space $\Theta$.

Diversity algorithms, also called {\it divergent search} (DS) algorithms, are policy search algorithms dedicated to covering a space of solutions as widely as possible. In particular, they can be used to find a target area in the absence of a reward signal. A common feature of these algorithms is that they define an {\it outcome space} $\mathcal{O}$ as a generally low-dimensional space that can characterize important properties of policy runs.
The target area in such policy search problems is generally defined in $\mathcal{O}$. Thus it is natural to consider that DS algorithms are performing search in that space and to define the selection operator in that space.

But a key issue in the policy search context is that one cannot directly sample in $\mathcal{O}$, as the mapping from outcomes to policy parameters reaching these outcomes is generally unknown. As a consequence, search in these DS algorithms considers the mapping between $\Theta$ and $\mathcal{O}$, which we call the $f: \Theta \rightarrow \mathcal{O}$ mapping hereafter, see \figurename~\ref{fig:mappings_diversity_algorithms2}.

The necessity to consider these two spaces results in key differences between MP and DS algorithms. In particular, while MP algorithms need to use a local expansion operator because they build a path to control a system from one configuration to another, DS algorithms rely on local expansions for different reasons.

Importantly, as it is not possible to sample directly in $\mathcal{O}$, the expansion operator must sample in $\Theta$. Since selection operates in $\mathcal{O}$ and expansion in $\Theta$ but from the selected sample, one must determine the $\theta \in \Theta$ corresponding to the selected $o \in \mathcal{O}$. This problem is easily solved by storing in the archive a pair consisting of a $\theta$ and the resulting outcome $o$ for each sample. For a selected $o$, a common approach for expansion is to simply apply a random mutation to the corresponding $\theta$.
For the selection operator, the NS and GEP algorithms respectively implement the two strategies described in Section~\ref{sec:sel_exp}. Their pseudo-codes are given in Appendix~\ref{sec:algo_DA}.

\subsubsection{Selection in NS}

Novelty Search considers two sets of points in $\mathcal{O}$: the population and the archive. Only the policies contained in the population may be selected. We explain later how these sets of points are constructed. 

Selection in NS can be performed using various selection operators. The uniform selection operator, the score proportionate selection operator, and the tournament-based operators are the most common ones \cite{cully_quality_2017}. In this paper we focus on the score proportionate selection operator biased toward more novel policies. 

The idea behind score proportionate selection is to construct a probability distribution according to the {\bf novelty scores} of the policies contained in the population. The novelty score $\mathcal{N}$ of a point $o\in\mathcal{O}$ is defined as the average distance to the k-nearest neighbors $(\mu_1, \dots, \mu_k) \in \mathcal{O}^k$ in the archive, $k$ being a hyper-parameter:

\begin{equation}
\label{eq:ns_nov}
    \mathcal{N} = \frac{1}{k}\sum_{i=1}^{k}dist(o,\mu_i).
\end{equation}

Given a population $\{(\theta_{i}, o_{i})\}_{i\in \{1,\dots,N\}}$ containing $N$ policies, the probability $p_{k}$ for policy $\theta_{k}$ to be selected is proportional to its novelty score:
\begin{equation}
\label{eq:ns_selec}
    p_k = \frac{\mathcal{N}_{k}}{\sum_{i=0}^{N}\mathcal{N}_{i}}.
\end{equation}

This is an instance of Strategy~\ref{strat:est} described in Section~\ref{sec:sel_exp} where the distance to neighbors is computed through the novelty score.

\subsubsection{Selection in GEP}
\label{sec:selec_gep}

The selection operator in GEP works as follows. First, the agent draws a random target outcome $o_{goal}$. The agent would like to find a set of policy parameters $\theta_{goal}$ producing $o_{goal}$. For that, it looks in the archive for the closest outcome $o_{sel}$ to $o_{goal}$, and it selects the policy parameters $\theta_{sel}$ which generated $o_{sel}$. This is clearly an instance of Strategy~\ref{strat:rrt}.

Since the GEP selection operator draws a random outcome and selects a policy corresponding to the closest outcome in the archive $\{(\theta_{i}, o_{i})\}_{i\in \{1,\dots,N\}}$, the probability $p_{k}$ for policy $\theta_k$ contained in the archive to be selected is proportional to the volume of the Voronoi cell of its outcome $o_{k}$, as explained for RRT in Section~\ref{sec:rrt}: 
\begin{equation}
\label{eq:gep_selec}
    p_k = \frac{volume(Voronoi\ cell\ o_{k})}{\sum_{i=0}^{N}volume(Voronoi\ cell\ o_{i})}.
\end{equation}

One can immediately see that the selection operator is exactly the same as in RRT, but acting in a different space.

\subsubsection{Filtering in NS}
\label{sec:filtering}

In addition to their selection operators, NS also differs from GEP by using two filtering mechanisms. 

\begin{figure}[ht!]
    \centering
    \begin{subfigure}[ht]{0.45\textwidth}
         \centering
         \includegraphics[width = \hsize]{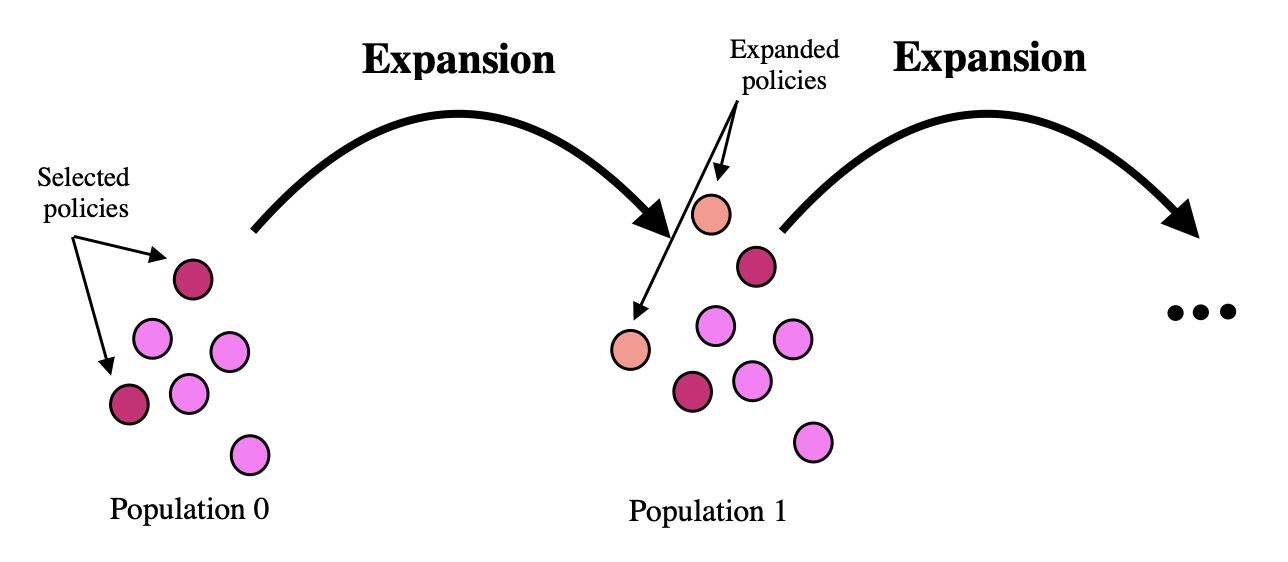}
         \caption{Goal-Exploration Process.}
        \label{fig:expansion_GEP}
    \end{subfigure}
    \hfill
    \begin{subfigure}[ht]{0.45\textwidth}
         \centering
         \includegraphics[width = \hsize]{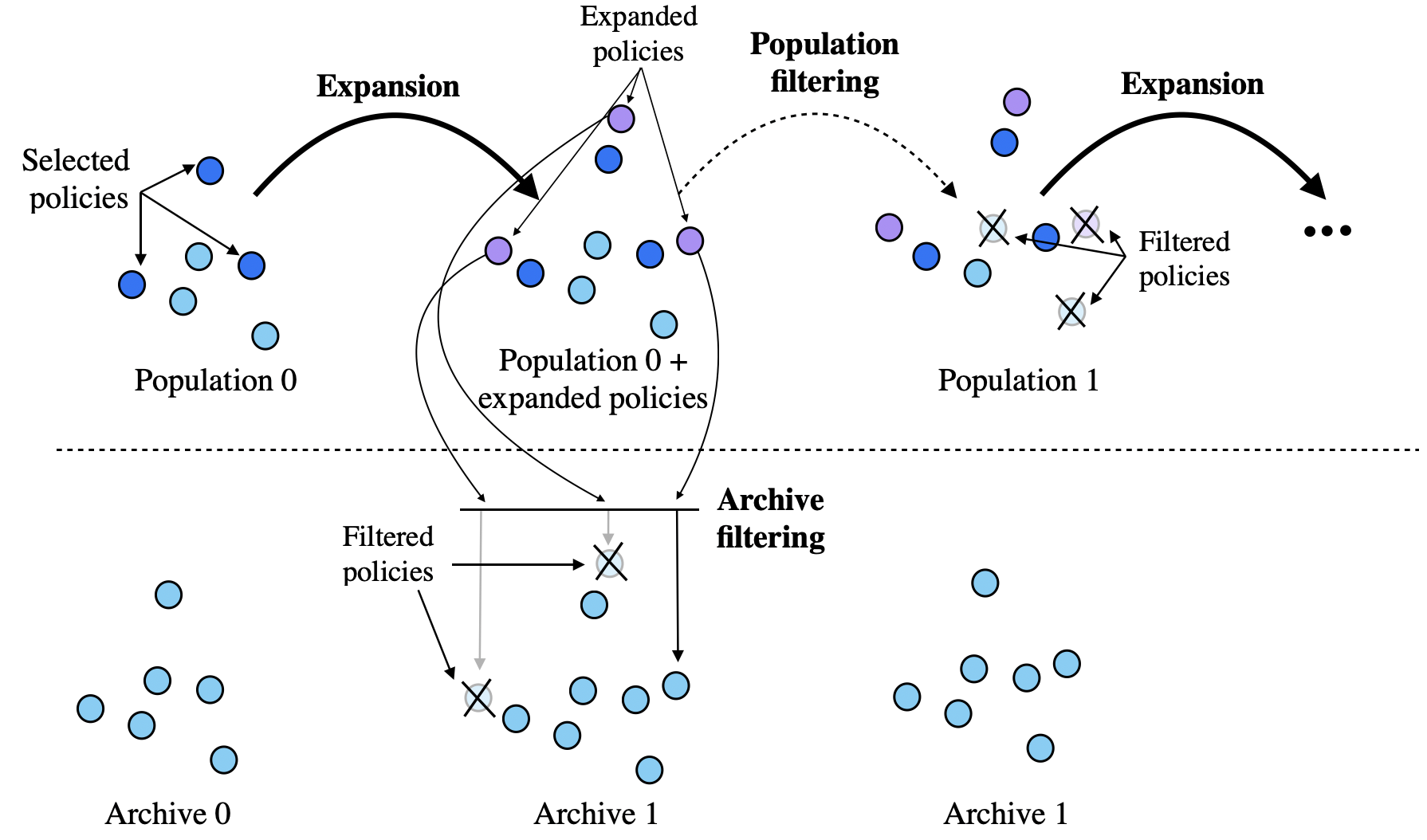}
         \caption{Novelty Search.}
        \label{fig:expansion_NS}
    \end{subfigure}
        \caption{Expansions in GEP and NS. In GEP, all expanded policies are added to the population. In NS, two filtering mechanisms are applied to the population and the archive.}
        \label{fig:expansion_GEP_NS}
\end{figure}

\begin{figure*}[ht!]
    \centering
    \includegraphics[width = \hsize]{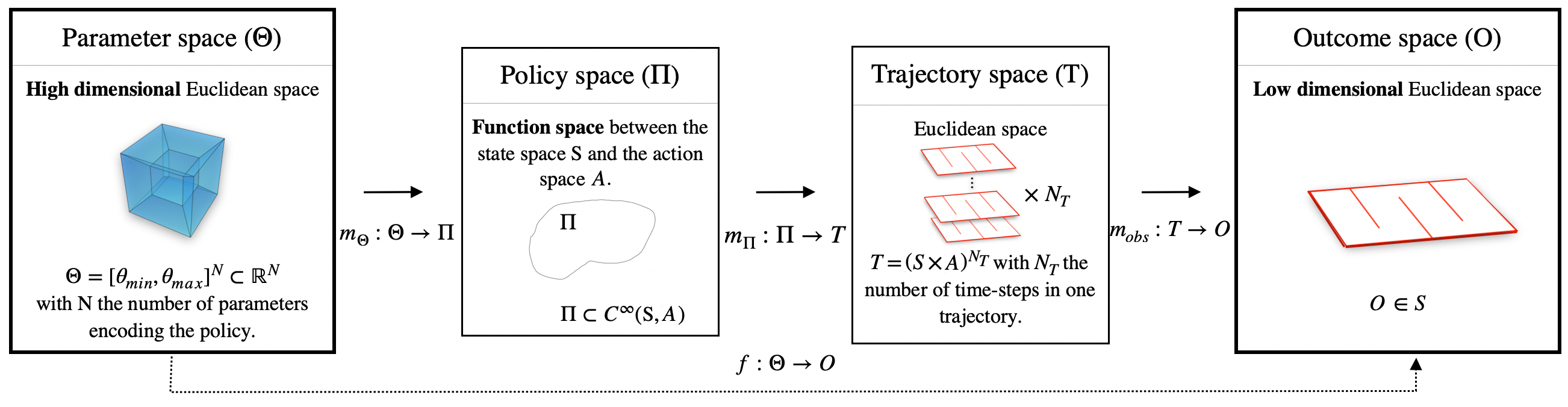}
    \caption{Description of the different spaces and mappings composing the $\Theta \rightarrow \mathcal{O}$ mapping in diversity search algorithms. Two intermediate spaces are considered: a policy space $\Pi$ and a trajectory space $T$. Three sub-mappings $m_{\Theta}, m_{\Pi}$ and $m_{obs}$ are also considered such that $f=m_{\Theta}\circ m_{\Pi}\circ m_{obs}$.}
    \label{fig:mappings_diversity_algorithms2}
\end{figure*}

\textbf{\textit{Filtering the population}}

The notion of population differs in GEP and NS. In GEP, the population gathers all policies since the first generation (see \figurename~\ref{fig:expansion_GEP}). At each iteration, all expanded policies are added to the population. In NS, the population is composed of a fixed size set of policies updated at each generation. As in GEP, it is initialized with random policies. However, after expanding the policies contained in the population, only the most novel policies contained in the set $\{$ new population + new offspring $\}$ are selected to construct the new population. This approach encourages the policies to move in the outcome space from one generation to another and thus promotes exploration \citep{doncieux2020}.
\par\vspace{2mm}
\textbf{\textit{Filtering the archive}}

Beyond the population, NS uses another set called the archive to keep track of the policies evolved in past generations. The archive is initialized with the random policies used to initialize the population. At each generation, after expanding the population, about $10\%$ policies are randomly sampled among the offspring and added to the archive, to keep the archive small. Indeed, the archive being used to compute the novelty score, keeping it small limits the cost of finding the k-nearest neighbors.

The archive in NS is only used to compute the novelty score of policies contained in the $\{\mbox{population + offspring}\}$ set. Policies from the archive are not added to the new population. If a policy contained in the $\{$ population + offspring $\}$ set is not selected for expansion, it is discarded and lost for future generations. 

\subsection{Similarities between MP and DS algorithms}
\label{sec:similarities_MP_DA}

It should now be obvious that, if we consider their most local expansion operators, NS shares similarities with EST and GEP with RRT. Indeed, the selection and expansion operators of the DS algorithms are closely related to the same operators of their MP counterpart. 

\subsubsection{Selection}

From the side of NS and EST, their selection operators measure how isolated a sample is by attributing a weight to each sample proportional to the inverse of the density of the archive in its neighborhood. The variants of EST and NS considered in this paper use the same weight computation based on the mean distance of samples to their k-nearest neighbor in the exploration tree (see \eqref{eq:est_weight}) or the archive (see \eqref{eq:ns_nov}). Therefore, nodes in EST and policies in NS share the same selection probability (see \eqref{eq:est_selec} and \eqref{eq:ns_selec}) in different spaces. 

Similarly, GEP and RRT also use a similar selection operator based on the volume of the Voronoi cells of the nodes/outcomes (see \eqref{eq:rrt_selec} and \eqref{eq:gep_selec}) in their respective space.

\subsubsection{Expansion}

The expansion operator used in MP highly depends on how much we know how to steer a system into a desired direction. But, in the unknown system case, a standard strategy consists in applying a single random control.
This strategy relies on a strong assumption about the dynamical system. In order to ensure that the algorithm is guaranteed to find a path between the starting configuration and the goal configuration given an infinite amount of selection-expansion iterations (i.e the algorithm is probabilistic complete  \citep{lavalle_planning_2006}), the system is assumed to be Lipschitz-continuous \citep{kleinbort_probabilistic_2018}. This assumption means that with enough expansions from the same node, a node should finally expand in the right direction. 

In DS algorithms, the standard expansion operator applies a random perturbation to the selected policy parameters, which has similarities with the use of random actions for local expansions in the MP context.

However, reasons for using a local expansion operator are different in the MP and DS contexts. In the MP context, one needs to locally control the system along a path from the current configuration to the target configuration. In GEP, a local random perturbation is applied to the selected policy hoping that, the corresponding outcome being close to the sampled goal, the perturbed policy will produce an outcome that is also close (and possibly closer) to this goal. One can see that the application of this selection-expansion strategy relies on the assumption of a smoothness property in the $f:\Theta \rightarrow \mathcal{O}$ mapping, i.e. that similar parameters yield similar outcomes.
In the case of NS, the reason for using a local expansion operator is less straightforward. It relies on the assumption that, if a policy resulted in an outcome in a low density region, a perturbed version of the policy should also result in an outcome exploring this low density region, and thus be potentially helpful in the search of new outcomes. Again, this is equivalent to assuming a smoothness property in the $f$ mapping.

\subsection{Expansions in DS are often non-local}
\label{sec:expansion_DS}

\begin{figure}[ht]
    \centering
   \begin{subfigure}{0.49\hsize}
         \centering
          \includegraphics[width = \hsize]{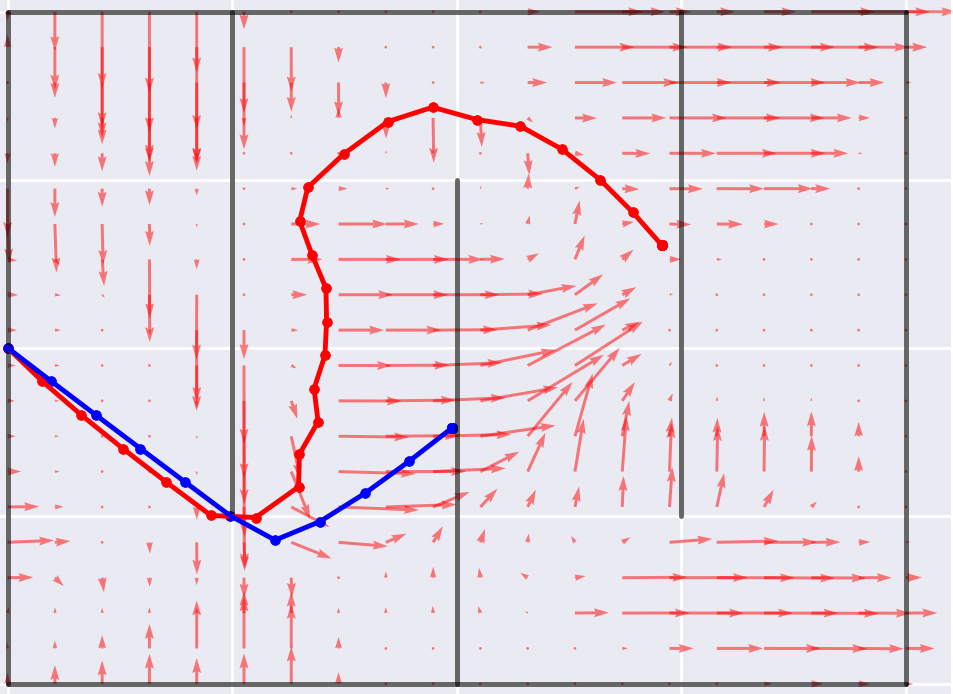}
    \caption{}
    \label{fig:non_local_f_theta2}
    \end{subfigure}
    \hfill
    \begin{subfigure}{0.49\hsize}
         \centering
         \includegraphics[width =\hsize]{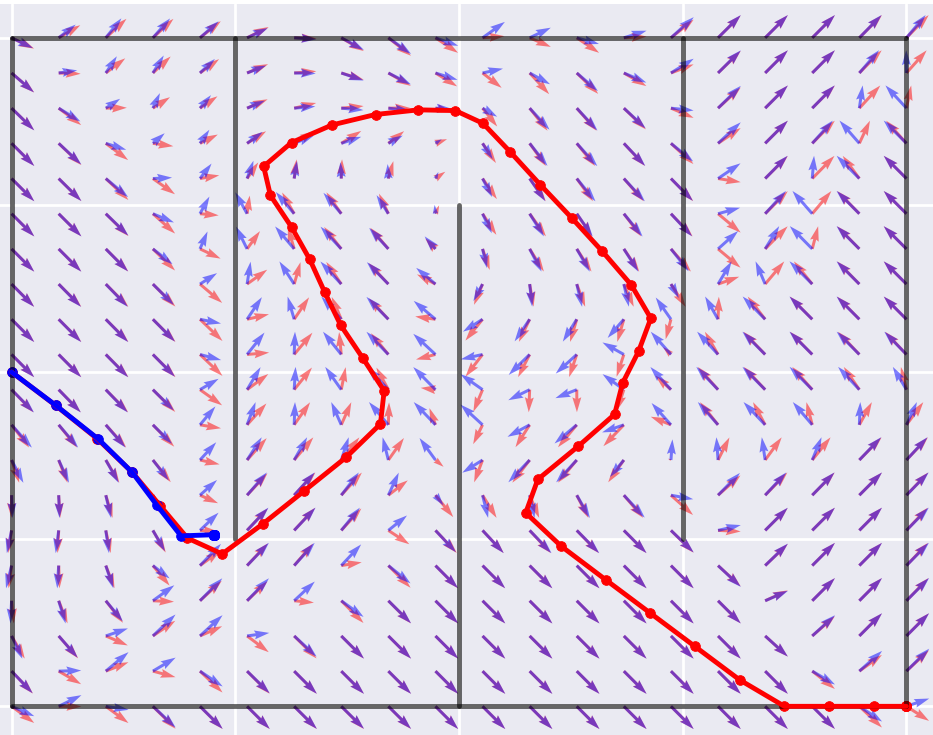}
    \caption{}
    \label{fig:non_local_f_pi2}
    \end{subfigure}
    \caption{(a) Lack of smoothness caused by the non-linearity of the neural network policy. The difference between two policies $\pi_\theta$ and $\pi_{\theta + \epsilon}$ is visualized as a vector field (in red). $\pi_\theta$ yields the red trajectory, and $\pi_{\theta+\epsilon}$ yields the blue trajectory. Although the parameters are close in $\Theta$, output differences are accumulated and lead to very different trajectories. (b) Lack of smoothness caused by the environment. The non-linearity or discontinuity of the environment (here at the extremity of the first wall) can cause similar actions to have dramatically different effects. This can lead to huge outcome differences for similar policy parameters.}
    
    \label{fig:expansion_DS}
\end{figure}

Even though DS have good reasons to use local expansions just like MP, expansions in DS are often non-local. Indeed, there are different sources of non-locality that can be identified by dissecting the $f:\Theta \rightarrow \mathcal{O}$ mapping and considering a policy space $\Pi$ and a trajectory space $T$ (see \figurename~\ref{fig:mappings_diversity_algorithms2} for the details of the sub-mappings).

The first source of non-locality originates from the $m_{\Theta}$ mapping relating $\Theta$ to policies, 
in particular when they are modeled as non-linear neural networks. Even though Multi-Layers Perceptrons (MLPs) are continuous functions, if the magnitude of the perturbation is too large, the expanded version of a policy may yield a very different policy. \figurename~\ref{fig:non_local_f_theta2} illustrates the consequences of a random mutation.

The second source of non-locality lies in the nature of outcomes, which, as mentioned in section \ref{sec:appli_div_search}, depend on policy runs, and therefore on trajectories. After selecting a pair $(\theta_{sel}, o_{sel}) \in \Theta \times O$, the expansion operator in DS perturbs $\theta_{sel}$ to obtain a new policy with parameters $\theta_{new} = \theta_{sel} + \delta \theta$ with $\delta \theta$ sampled from a spherical Gaussian distribution \cite{deep_neuroevo} or a more complex distribution \cite{Deb2014}.
The new policy $\pi_{\theta_{new}}$ yields trajectories defined by the equation

\begin{equation}
    \label{eq:expansion_DA}
    s(T) = s_{0} + \int_{0}^{T}\mathcal{D}_{sys+env}(s(t), {\pi_{\theta_{new}}(}s(t))dt
\end{equation}
\noindent
which integrates the dynamical system $\dot{s} = \mathcal{D}_{sys+env}(s, \pi_{\theta_{new}}(s))$ over the time interval $[0, T]$, where $\mathcal{D}_{sys+env}$ models the dynamics of the system in interaction with its environment and $s_{0}$ is the starting state of the rollout.
Even if the magnitude of the mutation is kept low enough for the expanded policies to be very close to the selected one, the numerous time steps of control may result in a large deviation between the trajectories obtained by the two policies via Equation~\eqref{eq:expansion_DA} as differences accumulate over time steps. These errors may be aggravated by discontinuous dynamical systems or environments $\mathcal{D}_{sys+env}$ and result in a non-smooth mapping $m_{\Pi}$ from policies to trajectories, and therefore from policies to outcomes. For instance, in maze environments, two close policies may yield very different trajectories if one trajectory gets blocked by a wall, as illustrated in \figurename~\ref{fig:non_local_f_pi2}.

\subsubsection{Preliminary conclusion: performance assumptions}

The similarities outlined above suggest that, if the expansion operators have similar properties, NS and GEP should share exploration abilities that are similar to those of EST and RRT respectively. However, for common $f:\Theta \rightarrow \mathcal{O}$ mappings, small perturbations in $\Theta$ may result in large changes in $\mathcal{O}$, and this lack of smoothness can result in a two situations.

\begin{itemize}
    \item 
    If the lack of smoothness of the $f:\Theta \rightarrow \mathcal{O}$ mapping is too serious, one could hypothesize that the use of local expansions in DS should bring no advantage compared to a random sampling of policy parameters.

    \item 
    Or, the $f:\Theta \rightarrow \mathcal{O}$ mapping could be smooth enough to let NS and GEP both outperform random sampling, but not smooth enough to inherit the search properties of EST and RRT.
\end{itemize}
\noindent
Below, we investigate these two possibilities experimentally.

\section{Experimental study}
\label{sec:experimental_study}

In this section, we experimentally study NS, GEP and a random search baseline using two environments with different smoothness properties to assess whether NS and GEP inherit from the properties of EST and RRT. The pseudo-codes of both MP and both DA algorithms are given in Appendix \ref{sec:algorithms}.

\subsection{Experimental setup}
\label{sec:setup}
\begin{figure}[ht]
    \centering
   \begin{subfigure}[ht]{0.49\hsize}
         \centering
        \includegraphics[width =  \hsize]{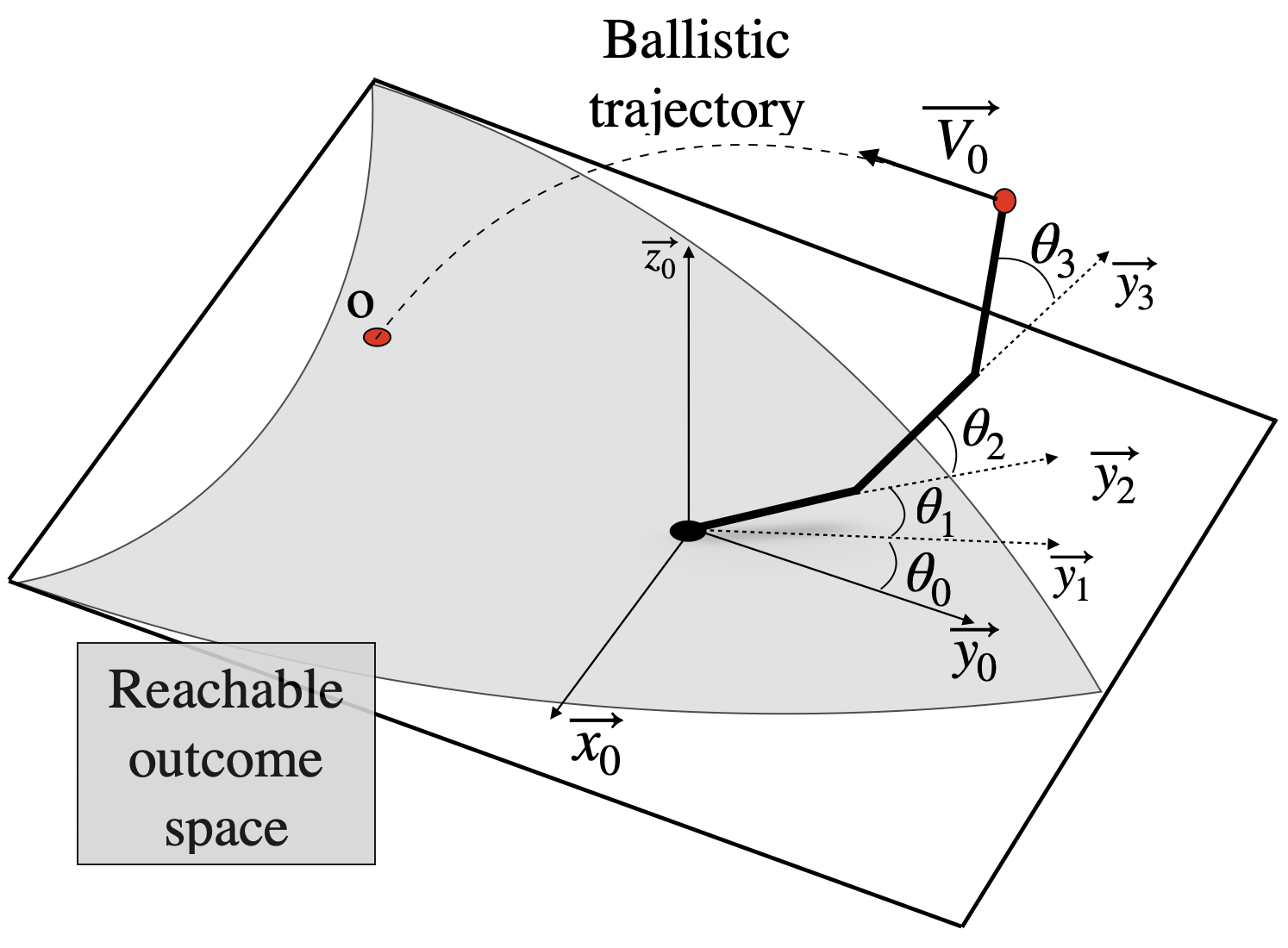}
    \caption{3D ballistic throw.}
    \label{fig:balistic_task}
    \end{subfigure}
    \hfill
    \begin{subfigure}[ht]{0.49\hsize}
         \centering
         \includegraphics[width = \hsize]{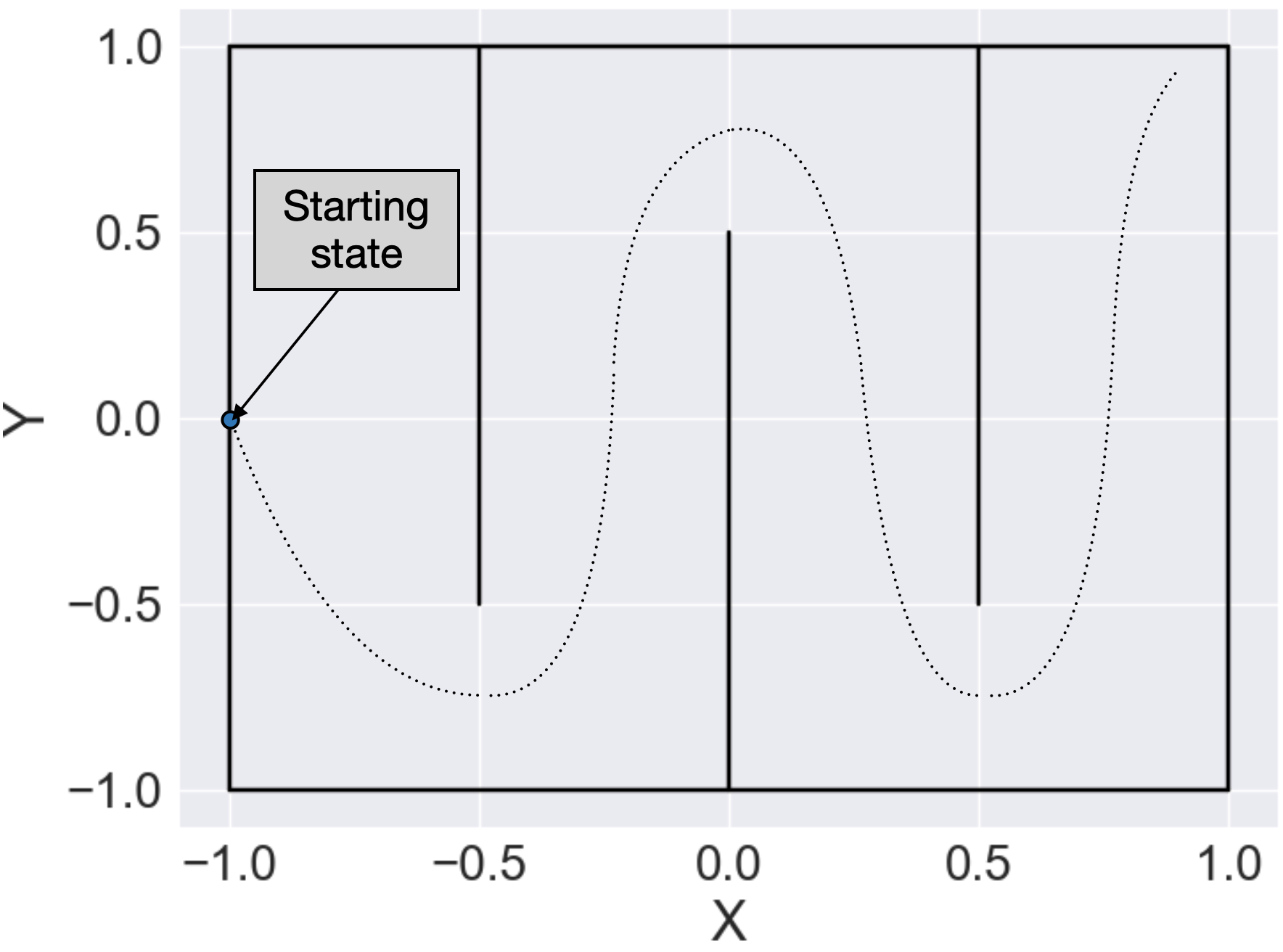}
         \caption{SimpleMaze.}
        \label{fig:simplemaze}
    \end{subfigure}
      \caption{Studied environments. (a) In 3D ballistic throw, an agent controls the angular speed $(\dot{\theta_{i}})_{i\in[0,3]}$ of a 4-joint 3D robot arm in order to throw a projectile. The outcome $o$ of the policy is the final position of the projectile. (b) In SimpleMaze, the agents start from a fixed position $(-1,0)$ and must reach the upper right corner.}
        \label{fig:envs}
\end{figure}

The experimental comparison is based on two environments: a ballistic task using a 4-DOF simulated robot arm and a maze environment called SimpleMaze, see \figurename~\ref{fig:envs}. In both environments, the state space is continuous and time is discrete. 

\begin{figure*}[ht!]
    \centering
    \begin{subfigure}[ht]{0.27\textwidth}
         \centering
         \includegraphics[width = \hsize]{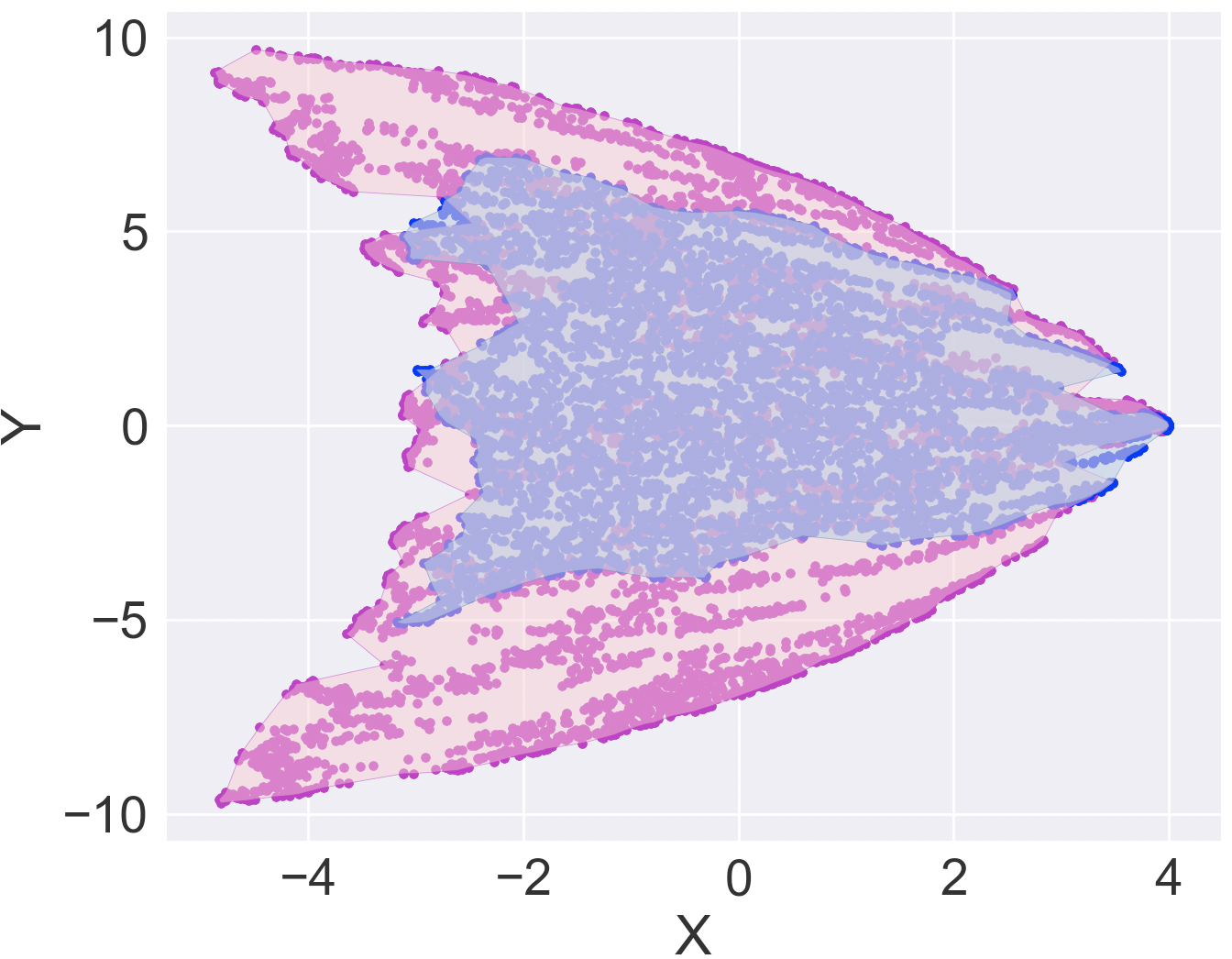}
         \caption{Concave hull 3D ballistic throw.}
        \label{fig:coverage_ballistic_3D}
    \end{subfigure}
    \hfill
    \begin{subfigure}[ht]{0.29\textwidth}
         \centering
         \includegraphics[width = \hsize]{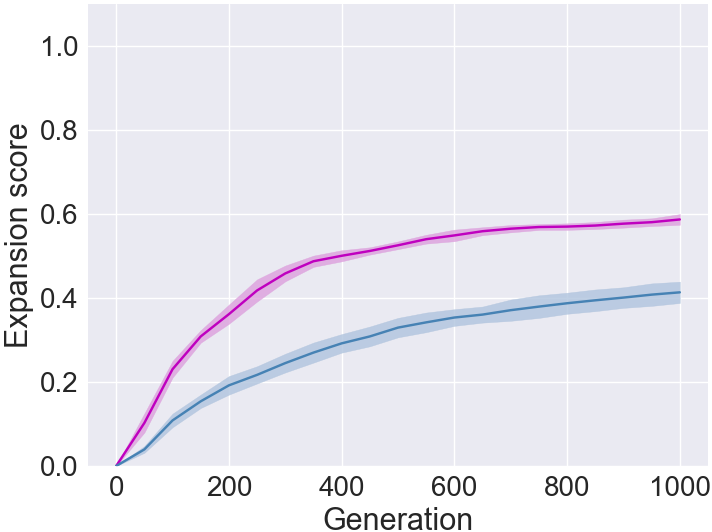}
         \caption{Expansion 3D ballistic throw.}
        \label{fig:expansion_score_ballistic}
    \end{subfigure}
    \hfill
    \begin{subfigure}[ht]{0.29\textwidth}
         \centering
         \includegraphics[width = \hsize]{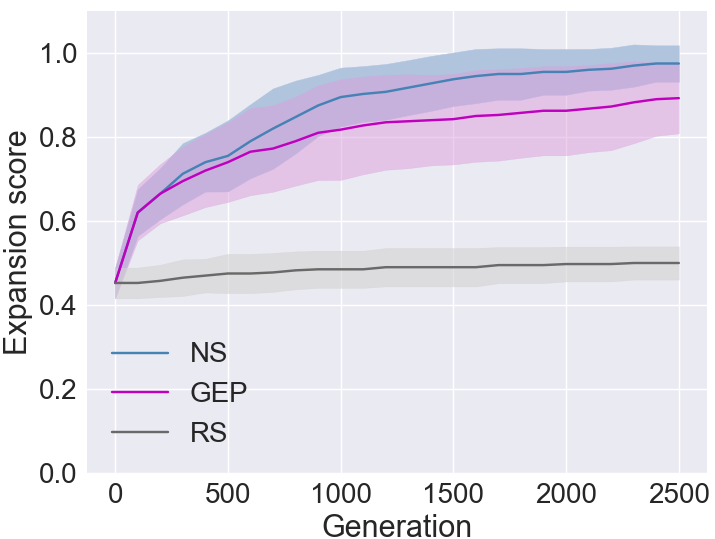}
         \caption{Expansion SimpleMaze.}
        \label{fig:expansion_large_simple}
    \end{subfigure}
        \caption{(a) Visualisation of the concave hull of the exploration trees of NS (blue) and GEP (pink) in 3D ballistic throw. (b) Expansion scores of GEP and NS in 3D ballistic throw. As the expansion operator is local, GEP inherits exploration properties from RRT and expands faster than NS. (c) Expansion scores of GEP, NS and a random search (RS) baseline in SimpleMaze. The non-local expansions result in NS exploring slightly faster than GEP. Both NS and GEP outperform RS.}
        \label{fig:results_hull}
\end{figure*}

\paragraph{3D ballistic throw}
\label{sec:ballisticarm}

The planar robot arm ballistic throw environment simulates the trajectory of a projectile thrown by a 3D 4-joint robot arm inspired from \cite{Cully2019}. The velocities $(\dot{\theta_{i}})_{i\in[0,3]} \in [-1 rad.s^{-1}, 1rad.s^{-1}]$ of the joints of the robot arm are controlled by an MLP. The throw is divided into acceleration-release phases. The acceleration phase is a single time step of control of the robot joints. After the acceleration phase, the end effector of the robot releases the projectile which then follows a ballistic trajectory. The outcome is the $(x,y)$ coordinates of impact. 

The expansion operator of DS algorithms truly achieves local expansions in this environment. First, by decreasing the magnitude of the polynomial mutations with $\eta = 2000$ \cite{Deb2014}, we make sure that any mutated policy is similar to the one from which it originated. Second, by reducing the acceleration phase to a single time step, we avoid the accumulation of differences between trajectories of the dynamical system controlled by the mutated policy and by its parent. 
This ensures the smoothness of the $m_{\Theta}$ mapping. Moreover, without obstacles, there is no discontinuity in the trajectory of the projectile, and the function $\mathcal{D}_{sys+env}$ (Equation~\eqref{eq:expansion_DA}) remains smooth under all circumstances.
Under these constraints, the $f : \Theta \rightarrow \mathcal{O}$ mapping is such that small perturbations in $\Theta$ yield small changes in $\mathcal{O}$. 

\paragraph{SimpleMaze}

We chose a Maze environment as it facilitates visualization of the exploration properties of the algorithms. 
A rollout lasts 50 time steps. The outcome corresponds to the final position of the agent at the end of the rollout. The agent starts from $(-1,0)$ and receives at each time step the position of $(x,y)\in [-1,1]\times[-1,1]$ at the current time step as input and outputs the next displacement $(dx,dy)\in [-0.1,0.1]\times[-0.1,0.1]$. The agent does not perceive the walls. 
The magnitude of the polynomial mutations ($\eta = 15$), the duration of a rollout as well as the discontinuities caused by the walls result in non-local expansions (as shown in Figure \ref{fig:expansion_DS}), with similar policy parameters leading possibly to very different outcomes.

\subsubsection{Metrics, hyper-parameters and technical details}
\label{sec:metrics}

In order to assess the expansion of a DS algorithm, we use the previously defined domain expansion metric. We split $\mathcal{O}$ into a $G_{expansion}\times G_{expansion}$ expansion grid. The expansion score corresponds to the number of expansion cells filled with at least one policy over the total number of expansion cells.
Implementations details are presented in Appendix~\ref{sec:implem} and hyper-parameters are summarized in Appendix~\ref{sec:hyperparams}.

\subsection{Results}
\subsubsection{Results on 3D ballistic throw}
\label{sec:ballistic_arm_results}

In this section, we verify that in this environment GEP and NS inherit the search properties of RRT and EST, and that, as expected, GEP explores $\mathcal{O}$ faster than NS.

Figure~\ref{fig:expansion_score_ballistic} presents the expansion score obtained by GEP and NS after 1000 generations. It confirms that GEP expands faster across $\mathcal{O}$ and converges quickly towards a maximum\footnote{The non-rectangular shape of $\mathcal{O}$ in the 3D ballistic throw environment makes some cells of the expansion grid unreachable, which explains why GEP eventually covers only about $\sim 60\%$ of $\mathcal{O}$.} that NS struggles to reach. 
These results are analogous to the performance of RRT and EST described above. The similarities between the selection operator of both DS algorithms and their MP counterpart enables similar exploration performance in the ballistic task, an environment that preserves the locality of expansions.

\subsubsection{Coverage visualization}
\label{sec:visu_cov}

\figurename~\ref{fig:coverage_ballistic_3D} presents the concave hull obtained by the exploration trees of GEP and NS after 500 generations in one run. We observe that NS spends numerous generations paving the center of the reachable outcome space ($|Y| < 5$) whereas GEP expands in all directions and quickly finds the limits of the reachable search space. 


\subsubsection{Results in SimpleMaze}
\label{sec:maze_results}

\begin{figure}[!ht]
    \centering
    \includegraphics[width = 0.6\hsize]{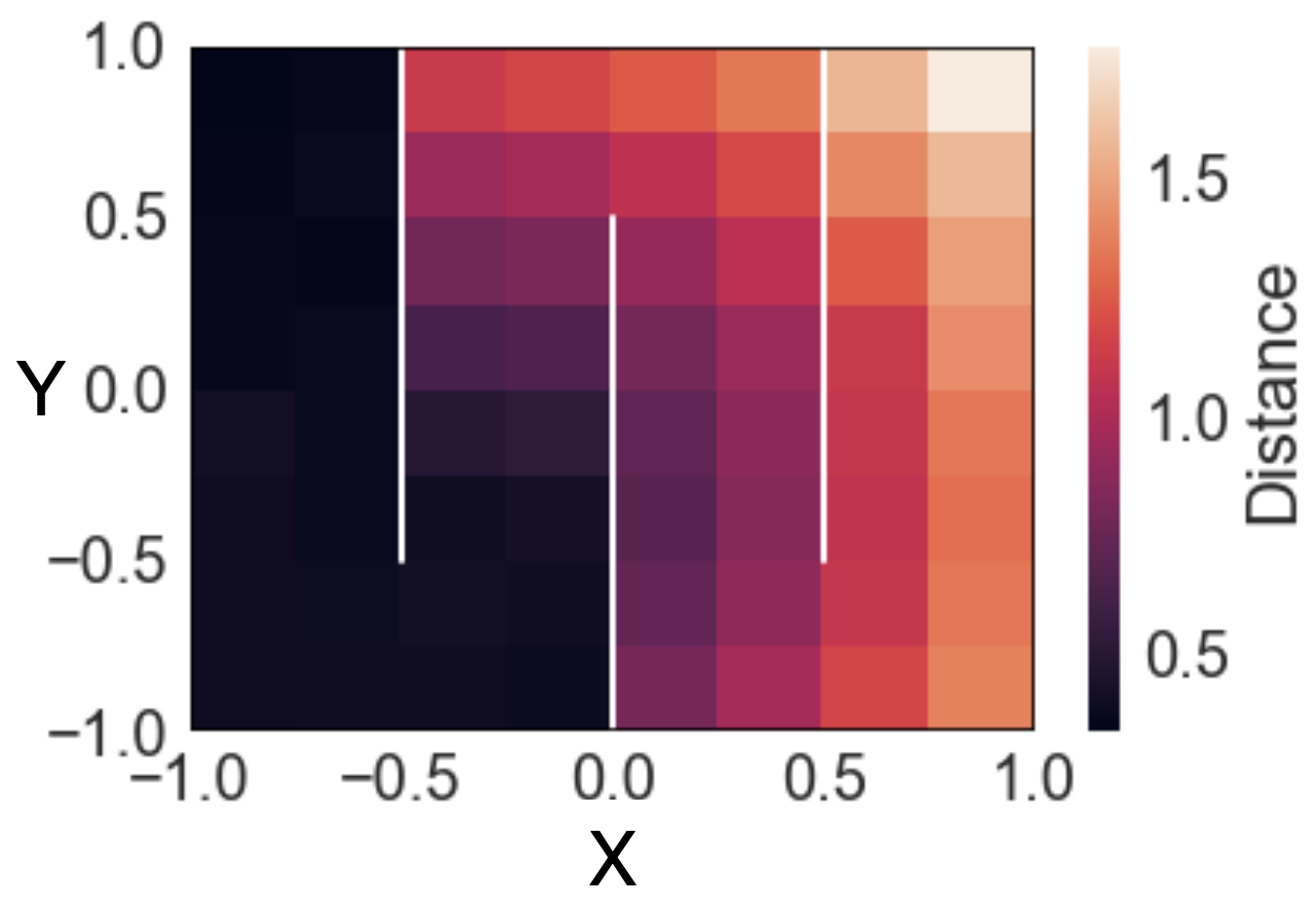}
    \caption{Mean distance between the outcome of a selected policy and the outcomes of its expanded versions depending on its position in the SimpleMaze. We observe that this distance increases as we progress through the maze i.e the expansion operator gets increasingly more non-local. The mean distances are computed for 100 expansions per selected policies and 200 selected policies per cell. Results are averaged over 10 runs of NS and 10 runs of GEP.}
    \label{fig:expansion_degradation}
\end{figure}

We previously argued that, if the expansion operator is local, NS and GEP have search properties that are similar to their MP counterparts. 
We also explained that the $f: \Theta \rightarrow \mathcal{O}$ mapping is not smooth in SimpleMaze, which results in non-local expansions. In this section we assess the consequences of non-local expansions by comparing the exploration performance of NS and GEP. We show that this non-locality results in loosing the properties inherited from RRT and EST. 

\figurename~\ref{fig:expansion_large_simple} presents the expansion and density scores obtained by NS, GEP and RS in Simple Maze. 
It shows that both NS and GEP outperform RS. Therefore, the $f:\Theta\mapsto O$ mapping is smooth enough for NS and GEP to benefit from their selection operator.

However, the performances of NS and GEP in SimpleMaze differ from their performances in the ballistic throw. NS and GEP perform similarly during the first 500 generations. Then, the expansion of NS accelerates and outperforms the expansion of GEP by achieving a full exploration of the maze after about 2500 generations while GEP generally fails to reach the end of the maze and only reaches an expansion score of 0.9 after the same number of generations. 

The difference in expansion rates after 500 generations arises from a progressive degradation of the locality of the expansion operator. As explained in Section~\ref{sec:expansion_DS}, because of the non-locality caused by the discontinuous environment due to walls and the non-linear neural network policy, expanding a policy which yields an outcome in the second corridor or beyond often results in a policy blocked by the first wall. \figurename~\ref{fig:expansion_degradation} illustrates this degradation by underlining the increasing distance between the outcome of a selected policy and the outcome of its expanded versions as the selected policy progresses through the maze. We see that the further we progress through the maze, the more distant the outcome of an expanded policy get from the outcome of a selected policy. In other words, the better the policy, the less local the expansion operator becomes.

\begin{figure}[ht!]
    \centering
    \captionsetup{justification=centering}
    \begin{subfigure}[ht]{0.49\hsize}
         \centering
         \captionsetup{justification=centering}
         \includegraphics[width = 1.1\hsize]{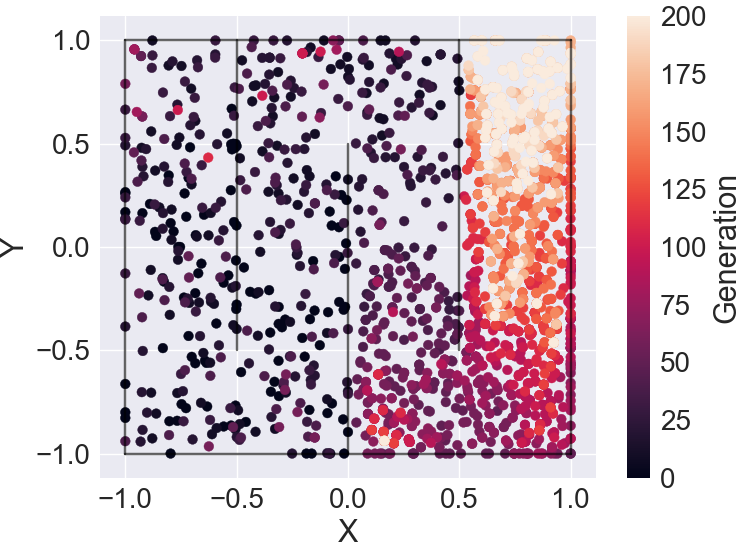}
         \caption{NS}
        \label{fig:selected_p_NS}
    \end{subfigure}
    \hfill
    \begin{subfigure}[ht]{0.49\hsize}
         \centering
         \captionsetup{justification=centering}
         \includegraphics[width = 1.1\hsize]{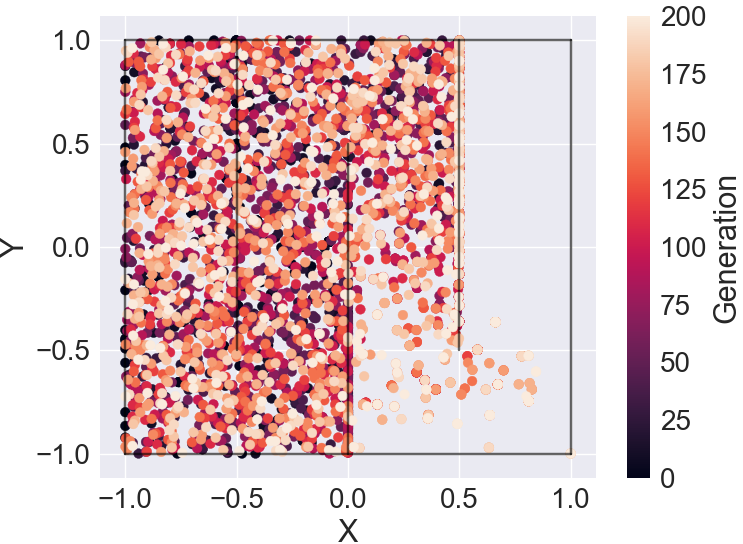}
         \caption{GEP}
        \label{fig:selected_p_GEP}
    \end{subfigure}
    \caption{History of the selected policies in SimpleMaze (2000 generations run).}
    \label{fig:selected_p}
\end{figure}

However, these better policies constitute promising stepping stones for further exploration \cite{Woolley2021}.
Therefore, it is important to select those most advanced policies to, eventually, discover policies yielding outcomes further in the maze. That is exactly what the filtering mechanisms of NS do. \figurename~\ref{fig:selected_p} presents the outcomes of the selected policies by NS and GEP through one run. In \figurename~\ref{fig:selected_p_NS}, the smooth change of colors shows that the selected policies by NS at a given generation correspond to the most advanced policies in the maze.

On the contrary, GEP does not integrate any filtering mechanisms. Therefore, GEP keeps sampling policies blocked in the already well-explored areas of the maze (see \figurename~\ref{fig:selected_p_GEP}). 
Thus, GEP requires more generations to deal with the degraded expansion operator which results in a slower increase of the expansion score.

These results show that, in an environment where the $f:\Theta \rightarrow \mathcal{O}$ mapping lacks smoothness, the properties inherited from RRT and EST are lost, which means that the selection-expansion mechanisms do not behave as originally intended. Even though both NS and GEP outperform random sampling of policies, additional heuristics such as filtering mechanisms must be exploited to overcome difficult expansions with degraded expansion operators.


\section*{Discussion \& Conclusion}
\label{sec:discussion}

In this article, we presented a comparison between two divergent search algorithms: GEP and NS. We started by presenting a unifying framework called selection-expansion which draws a parallel between both algorithms and two Motion Planning algorithms, EST and RRT.

An experimental study showed that in an environment like the 3D ballistic throw where the $f: \Theta \rightarrow \mathcal{O}$ mapping is smooth enough, GEP and NS inherit the exploration properties from their Motion Planning counterpart. In that case, GEP explores faster the environment than NS.

By contrast, maze results show that, even though GEP and NS share common selection-expansion properties with RRT and EST, they do not share the same exploration abilities if the expansion operator is not local. In such situations, the experimental study showed that NS outperforms GEP by using efficient filtering mechanisms.

This work opens up the question of restoring locality in complex environments where the expansion operator is non-local. 
In \cite{lehman_safe_2018}, the authors partially restore locality using safe mutations. However, safe mutations only tackle one source of non-localities coming from the non-linear nature of neural network policies. Discontinuities and non-linearities of the dynamical system like the walls in SimpleMaze are still an important source of non-locality in the expansion operators.
Based on this observation, the main research direction for future work should be to search for a generic way to restore a form of locality in environments where the expansion operator is not local. 

Furthermore, the results in SimpleMaze highlight the benefits of the filtering mechanism of NS when the algorithm is stuck in an exploration bottleneck. It could be interesting to search for a variant of GEP equipped with compatible filtering mechanisms. 

\section*{Acknowledgements}

This work was partially supported by the French National Research Agency (ANR), Project ANR-18-CE33-0005 HUSKI.


\begin{thebibliography}{29}
\providecommand{\natexlab}[1]{#1}
\providecommand{\url}[1]{\texttt{#1}}
\providecommand{\urlprefix}{URL }
\expandafter\ifx\csname urlstyle\endcsname\relax
  \providecommand{\doi}[1]{doi:\discretionary{}{}{}#1}\else
  \providecommand{\doi}{doi:\discretionary{}{}{}\begingroup
  \urlstyle{rm}\Url}\fi

\bibitem[{Akkaya et~al.(2019)Akkaya, Andrychowicz, Chociej, Litwin, McGrew,
  Petron, Paino, Plappert, Powell, Ribas et~al.}]{akkaya2019solving}
Akkaya, I.; Andrychowicz, M.; Chociej, M.; Litwin, M.; McGrew, B.; Petron, A.;
  Paino, A.; Plappert, M.; Powell, G.; Ribas, R.; et~al. 2019.
\newblock Solving Rubik's Cube with a Robot Hand.
\newblock \emph{arXiv preprint arXiv:1910.07113} .

\bibitem[{Andrychowicz et~al.(2018)Andrychowicz, Baker, Chociej, Jozefowicz,
  McGrew, Pachocki, Petron, Plappert, Powell, Ray
  et~al.}]{andrychowicz2018learning}
Andrychowicz, M.; Baker, B.; Chociej, M.; Jozefowicz, R.; McGrew, B.; Pachocki,
  J.; Petron, A.; Plappert, M.; Powell, G.; Ray, A.; et~al. 2018.
\newblock Learning dexterous in-hand manipulation.
\newblock \emph{arXiv preprint arXiv:1808.00177} .

\bibitem[{Bentley(1975)}]{Bentley1975}
Bentley, J.~L. 1975.
\newblock Multidimensional Binary Search Trees Used for Associative Searching.
\newblock \emph{Commun. ACM} 18(9): 509–517.
\newblock ISSN 0001-0782.

\bibitem[{Benureau and Oudeyer(2016)}]{benureau_behavioral_2016}
Benureau, F.; and Oudeyer, P.-Y. 2016.
\newblock Behavioral {Diversity} {Generation} in {Autonomous} {Exploration}
  through {Reuse} of {Past} {Experience}.
\newblock \emph{Frontiers in Robotics and AI} 3: 1--2.
\newblock ISSN 2296-9144.

\bibitem[{Cideron et~al.(2020)Cideron, Pierrot, Perrin, Beguir, and
  Sigaud}]{cideron2020qdrl}
Cideron, G.; Pierrot, T.; Perrin, N.; Beguir, K.; and Sigaud, O. 2020.
\newblock QD-RL: Efficient Mixing of Quality and Diversity in Reinforcement
  Learning.

\bibitem[{Colas, Sigaud, and Oudeyer(2018)}]{colas2016}
Colas, C.; Sigaud, O.; and Oudeyer, P. 2018.
\newblock {GEP-PG:} Decoupling Exploration and Exploitation in Deep
  Reinforcement Learning Algorithms.
\newblock \emph{CoRR} abs/1802.05054.

\bibitem[{Cully(2019)}]{Cully2019}
Cully, A. 2019.
\newblock Autonomous skill discovery with Quality-Diversity and Unsupervised
  Descriptors.
\newblock \emph{CoRR} abs/1905.11874.

\bibitem[{Cully and Demiris(2018)}]{cully_quality_2017}
Cully, A.; and Demiris, Y. 2018.
\newblock Quality and Diversity Optimization: A Unifying Modular Framework.
\newblock \emph{IEEE Transactions on Evolutionary Computation} 22(2): 245--259.

\bibitem[{Deb and Deb(2014)}]{Deb2014}
Deb, K.; and Deb, D. 2014.
\newblock Analysing mutation schemes for real-parameter genetic algorithms.
\newblock \emph{International Journal of Artificial Intelligence and Soft
  Computing} 4: 1--28.

\bibitem[{Doncieux, Laflaquière, and Coninx(2019)}]{doncieux_novelty_2019}
Doncieux, S.; Laflaquière, A.; and Coninx, A. 2019.
\newblock Novelty search: a theoretical perspective.
\newblock In \emph{Proceedings of the {Genetic} and {Evolutionary}
  {Computation} {Conference}}, 99--106. Prague Czech Republic: ACM.
\newblock ISBN 978-1-4503-6111-8.

\bibitem[{Doncieux et~al.(2020)Doncieux, Paolo, Laflaqui\`{e}re, and
  Coninx}]{doncieux2020}
Doncieux, S.; Paolo, G.; Laflaqui\`{e}re, A.; and Coninx, A. 2020.
\newblock Novelty Search Makes Evolvability Inevitable.
\newblock In \emph{Proceedings of the 2020 Genetic and Evolutionary Computation
  Conference}, GECCO ’20, 85–93. New York, NY, USA: Association for
  Computing Machinery.
\newblock ISBN 9781450371285.

\bibitem[{Ecoffet et~al.(2019)Ecoffet, Huizinga, Lehman, Stanley, and
  Clune}]{ecoffet2019goexplore}
Ecoffet, A.; Huizinga, J.; Lehman, J.; Stanley, K.~O.; and Clune, J. 2019.
\newblock Go-Explore: a New Approach for Hard-Exploration Problems.

\bibitem[{Forestier(2019)}]{forestier_these_2019}
Forestier, S. 2019.
\newblock \emph{Intrinsically Motivated Goal Exploration in Child Development
  and Artificial Intelligence: Learning and Development of Speech and Tool
  Use}.
\newblock Ph.D. thesis, U. Bordeaux.

\bibitem[{Forestier and Oudeyer(2016)}]{forestier_modular_2016}
Forestier, S.; and Oudeyer, P.-Y. 2016.
\newblock Modular active curiosity-driven discovery of tool use.
\newblock In \emph{2016 {IEEE}/{RSJ} {International} {Conference} on
  {Intelligent} {Robots} and {Systems} ({IROS})}, 3965--3972. Daejeon, South
  Korea: IEEE.
\newblock ISBN 978-1-5090-3762-9.

\bibitem[{Gu et~al.(2016)Gu, Holly, Lillicrap, and Levine}]{DRL_rob}
Gu, S.; Holly, E.; Lillicrap, T.; and Levine, S. 2016.
\newblock Deep {Reinforcement} {Learning} for {Robotic} {Manipulation} with
  {Asynchronous} {Off}-{Policy} {Updates}.
\newblock ArXiv: 1610.00633.

\bibitem[{Heess et~al.(2017)Heess, TB, Sriram, Lemmon, Merel, Wayne, Tassa,
  Erez, Wang, Eslami, Riedmiller, and Silver}]{DRL_mujoco}
Heess, N.; TB, D.; Sriram, S.; Lemmon, J.; Merel, J.; Wayne, G.; Tassa, Y.;
  Erez, T.; Wang, Z.; Eslami, S. M.~A.; Riedmiller, M.~A.; and Silver, D. 2017.
\newblock Emergence of Locomotion Behaviours in Rich Environments.
\newblock \emph{CoRR} abs/1707.02286.

\bibitem[{{Hsu}, {Latombe}, and {Motwani}(1997)}]{LatombePath}
{Hsu}, D.; {Latombe}, J.~.; and {Motwani}, R. 1997.
\newblock Path planning in expansive configuration spaces.
\newblock In \emph{Proceedings of International Conference on Robotics and
  Automation}, volume~3, 2719--2726 vol.3.

\bibitem[{Kleinbort et~al.(2018)Kleinbort, Solovey, Littlefield, Bekris, and
  Halperin}]{kleinbort_probabilistic_2018}
Kleinbort, M.; Solovey, K.; Littlefield, Z.; Bekris, K.~E.; and Halperin, D.
  2018.
\newblock Probabilistic completeness of {RRT} for geometric and kinodynamic
  planning with forward propagation.
\newblock \emph{arXiv:1809.07051 [cs]} ArXiv: 1809.07051.

\bibitem[{Lavalle(1998)}]{Lavalle98rapidly-exploringrandom}
Lavalle, S.~M. 1998.
\newblock Rapidly-Exploring Random Trees: A New Tool for Path Planning.
\newblock Technical Report 98-11, Computer Science Department, Iowa State
  University.

\bibitem[{LaValle(2006)}]{lavalle_planning_2006}
LaValle, S.~M. 2006.
\newblock \emph{Planning {Algorithms}}.
\newblock Cambridge: Cambridge University Press.
\newblock ISBN 978-0-511-54687-7 978-0-521-86205-9.

\bibitem[{Lehman et~al.(2018)Lehman, Chen, Clune, and
  Stanley}]{lehman_safe_2018}
Lehman, J.; Chen, J.; Clune, J.; and Stanley, K.~O. 2018.
\newblock Safe {Mutations} for {Deep} and {Recurrent} {Neural} {Networks}
  through {Output} {Gradients}.
\newblock \emph{arXiv:1712.06563 [cs]} ArXiv: 1712.06563.

\bibitem[{Lehman and Stanley(2011)}]{lehman_abandoning_2011}
Lehman, J.; and Stanley, K.~O. 2011.
\newblock Abandoning {Objectives}: {Evolution} {Through} the {Search} for
  {Novelty} {Alone}.
\newblock \emph{Evolutionary Computation} 19(2): 189--223.
\newblock ISSN 1063-6560, 1530-9304.

\bibitem[{Matheron, Perrin, and Sigaud(2019)}]{matheron2019problem}
Matheron, G.; Perrin, N.; and Sigaud, O. 2019.
\newblock The problem with {DDPG}: understanding failures in deterministic
  environments with sparse rewards.
\newblock \emph{arXiv preprint arXiv:1911.11679} .

\bibitem[{Matheron, Perrin, and Sigaud(2020)}]{matheron2020pbcs}
Matheron, G.; Perrin, N.; and Sigaud, O. 2020.
\newblock PBCS : Efficient Exploration and Exploitation Using a Synergy between
  Reinforcement Learning and Motion Planning.

\bibitem[{Silver et~al.(2016)Silver, Huang, Maddison, Guez, Sifre, Van
  Den~Driessche, Schrittwieser, Antonoglou, Panneershelvam, Lanctot
  et~al.}]{silver2016mastering}
Silver, D.; Huang, A.; Maddison, C.~J.; Guez, A.; Sifre, L.; Van Den~Driessche,
  G.; Schrittwieser, J.; Antonoglou, I.; Panneershelvam, V.; Lanctot, M.;
  et~al. 2016.
\newblock Mastering the game of Go with deep neural networks and tree search.
\newblock \emph{Nature} 529(7587): 484--489.

\bibitem[{Silver et~al.(2017)Silver, Schrittwieser, Simonyan, Antonoglou,
  Huang, Guez, Hubert, Baker, Lai, Bolton et~al.}]{silver2017mastering}
Silver, D.; Schrittwieser, J.; Simonyan, K.; Antonoglou, I.; Huang, A.; Guez,
  A.; Hubert, T.; Baker, L.; Lai, M.; Bolton, A.; et~al. 2017.
\newblock Mastering the game of Go without human knowledge.
\newblock \emph{Nature} 550(7676): 354--359.

\bibitem[{Such et~al.(2017)Such, Madhavan, Conti, Lehman, Stanley, and
  Clune}]{deep_neuroevo}
Such, F.~P.; Madhavan, V.; Conti, E.; Lehman, J.; Stanley, K.~O.; and Clune, J.
  2017.
\newblock Deep Neuroevolution: Genetic Algorithms Are a Competitive Alternative
  for Training Deep Neural Networks for Reinforcement Learning.
\newblock \emph{CoRR} abs/1712.06567: 1--2.

\bibitem[{Vinyals et~al.(2019)Vinyals, Babuschkin, Czarnecki, Mathieu, Dudzik,
  Chung, Choi, Powell, Ewalds, Georgiev et~al.}]{vinyals2019grandmaster}
Vinyals, O.; Babuschkin, I.; Czarnecki, W.~M.; Mathieu, M.; Dudzik, A.; Chung,
  J.; Choi, D.~H.; Powell, R.; Ewalds, T.; Georgiev, P.; et~al. 2019.
\newblock Grandmaster level in StarCraft II using multi-agent reinforcement
  learning.
\newblock \emph{Nature} 575(7782): 350--354.

\bibitem[{Woolley and Stanley(2012)}]{Woolley2021}
Woolley, B.~G.; and Stanley, K.~O. 2012.
\newblock Exploring Promising Stepping Stones by Combining Novelty Search with
  Interactive Evolution.
\newblock \emph{CoRR} abs/1207.6682.

\end{thebibliography}

\newpage
\clearpage
\appendix

\section{Implementation details}
\label{sec:implem}
In both GEP and NS, a KD-tree \citep{Bentley1975} is used to accelerate the nearest-neighbor computations. New policies are added at each generation. NS keeps the cost of reconstructing the KD-tree low via the filtering mechanism of its archive. However, in the vanilla version of GEP, every expanded policies are added to the tree. As a result, the size of the tree surges and reconstructing the kd-tree becomes computationally very expensive. To avoid reconstructing the tree at each iteration, a second small KD-tree keeps track of the most recent policies and is reconstructed at every generation. The main KD-tree is updated only every $N_{update}$ after transferring the content of the small KD-tree.

\section{Hyper-parameters}
\label{sec:hyperparams}
\begin{table}[ht]
\centering
\begin{tabular}{|p{0.2\hsize}||p{0.7\hsize}|}
 \hline
 \multicolumn{2}{|c|}{Symbols} \\
 \hline
 $N_{sel/exp}$ & Number of selection/expansion,\\
 \hline 
 $R_{neigh}$ & Radius of the neighborhood considered for weight computation (EST),\\
 \hline
 $s_{0}$ & Position of the starting point,\\
 \hline 
 $N_{samples}$& Number of samples per expansion (EST),\\
 \hline
 $N_{timestep}$ & Length of the trajectory,\\
 \hline
 $N_{selection}$ & Number of selected policies,\\
 \hline
 $N_{layers}$ & Number of layers of the MLP,\\
 \hline
 $N_{neurons}$ & Number of neurons per layer,\\
 \hline
 $N_{inputs}$ & Number of inputs of the policy,\\
 \hline
 $N_{outputs}$ & Number of outputs of the policy,\\
 \hline
 $N_{offspring}$ & Number of offspring per selected policy\\
 \hline
 $N_{generation}$ & Number of generation\\
 \hline
 $p_{expansion}$ & Probability for a selected policy to be expanded,\\
 \hline
 $p_{mutation}$ & Probability for a weight in a selected policy to be mutated,\\
 \hline
 $\eta$ & Parameter for the polynomial mutation\\
 \hline
 $k$ & Number of nearest neighbors considered for novelty computation (NS),\\
 \hline 
 $G_{expansion}$ & Grid size (expansion),\\
 \hline
 $G_{density}$ & Grid size (density),\\
 \hline
 $N_{filter\ archive}$ & Number of policies added to the archive per generation (GEP),\\
 \hline
\end{tabular}
\caption{List of symbols.}
\label{table:symbols}
\end{table}

\begin{table}[ht]
\centering
\begin{tabular}{|p{3cm}||l|l|l|}
 \hline
 \multicolumn{3}{|c|}{Hyper-parameters} \\
 \hline
 Parameter & RRT & EST\\
 \hline
 $N_{sel/exp}$ & 1000    & 1000 \\
 \hline 
 $R_{neigh}$ & None    & 0.2 \\
 \hline
 $s_{0}$ & (-1,0)   & (-1,0) \\
 \hline 
 $N_{samples}$ & None & 10 \\
 \hline 
 grid size (expansion) & 4$\times$4  & 4$\times$4\\
 \hline
\end{tabular}
\caption{Hyper-parameters used in the Motion Planning algorithms in the Simple Maze.}
\label{table:params_MP}
\end{table}

\begin{table}[hb]
\centering
\begin{tabular}{|p{3cm}||l|l|}
 \hline
 \multicolumn{3}{|c|}{Hyper-parameters} \\
 \hline
 Parameter & Ballistic & Simple Maze \\
 \hline
 $N_{timestep}$ & 1    & 50\\
 \hline
 $N_{selection}$ & 1    & 100\\
 \hline
 $N_{layers}$ & 2 & 2 \\
 \hline
 $N_{neurons}$ &50 & 50\\
 \hline
 $N_{inputs}$ & 5 & 2\\
 \hline
 $N_{outputs}$ & 5 & 2\\
 \hline
 $N_{offspring}$ &   2  & 2\\
 \hline
 $N_{generation}$ & 500 & 1000\\
 \hline
 $p_{expansion}$ & 1. & 1.\\
 \hline
 $p_{mutation}$ & 0.1 & 0.1\\
 \hline
 $\eta$ (polynomial 
 mutations)& 2000  & 15\\
 \hline
 $k$ & 15  & 15\\
 \hline
 $G_{expansion}$ & 10  & 4\\
 \hline
 $N_{filter\ archive}$ (NS) & 10 & 6\\
 \hline
 policy type & MLP & MLP \\
 \hline
\end{tabular}
\caption{Hyper-parameters used in the Diversity algorithms in the different experiments.}
\label{table:params_DP}
\end{table}

\onecolumn
\section{Algorithms}
\label{sec:algorithms}

\subsection{Motion Planning algorithms}
\label{sec:algo_MP}

\begin{algorithm}[ht]
\caption{Rapidly-Exploring Random Trees}
\begin{algorithmic}[1]    
    \State Initialize exploration tree:
    \State $T \leftarrow s_{0}$ 
    \While{$iteration < N_{sel/exp}$}
        \State $s_{samp} \leftarrow random\_config()$ \Comment{selection operator}
        \State $s_{sel} \leftarrow nearest\_neighbor(T,s_{samp})$
        \State $u_{rand} \leftarrow random\_control()$ \Comment{random action}
        \State $s_{new} \leftarrow expand(s_{sel}, u_{rand})$ \Comment{expansion operator}
        \State $T \leftarrow T.update((s_{sel}, s_{new}, u_{rand}))$ \Comment{update search tree}
    \EndWhile
\end{algorithmic}
\end{algorithm}

\begin{algorithm}[ht!]
\caption{Expansive Spaces Trees (NS variant)}
\begin{algorithmic}[1]    
    \State Initialize exploration tree:
    \State $T \leftarrow s_{0}$ 
    \While{$iteration < N_{sel/exp}$}
        \State $N \leftarrow compute\_weight(T)$ \Comment{compute weights (novelty-like)}
        \State $s_{sel} \leftarrow select(T,N)$ \Comment{selection operator (weight proportionate)}
        \State $u_{rand} \leftarrow random\_control()$ \Comment{random action}
        \State $s_{new} \leftarrow expand(s_{sel}, u_{rand})$ \Comment{expansion operator}
        \State $T \leftarrow T.update((s_{sel}, s_{new}, u_{rand}))$ \Comment{update search tree}
    \EndWhile
\end{algorithmic}
\end{algorithm}

\subsection{Diversity algorithms}
\label{sec:algo_DA}

\begin{algorithm}[ht!]
\caption{Vanilla Goal Exploration Process (Selection-Expansion variant)}
\begin{algorithmic}[1]    
    \State Initialize population:
    \State $P \leftarrow$ init\_population() 
    \While{$generation < N_{generation}$}
        \For{$i=1:N_{selection}$}
            
            \State $o_{goal} \leftarrow random\_outcome()$ \Comment{selection operator}
            \State $(\theta_{sel},o_{sel}) \leftarrow nearest\_neighbor(P,o_{goal})$
            \State $\theta_{new} \leftarrow expand(\theta_{sel})$ \Comment{expansion operator}
            \State $o_{new} \leftarrow evaluate(\theta_{new})$ \Comment{compute outcome}
            \State $P \leftarrow P + [(\theta_{new}, o_{new})]$
        \EndFor
    \EndWhile
\end{algorithmic}
\end{algorithm}

\begin{algorithm}[ht!]
\caption{Novelty Search (Selection-Expansion variant)}
\begin{algorithmic}[1]    
    \State Initialize population, expanded policies \& archive:
    \State $P \leftarrow$ init\_population() 
    \State $M \leftarrow [] $
    \State $A \leftarrow []$ 
    \While{$generation < N_{generation}$}
        \State $N \leftarrow novelty(\{P+M\}, {A+P})$ \Comment{compute novelty scores}
        \State $P' \leftarrow []$ 
        \For{$i=1:N_{selection}$} \Comment{population filtering}
            \State $P'\leftarrow P' + [select(\{P+M\}, N)]$ \Comment{selection operator (novelty proportionate)}
        \EndFor
        \State $P \leftarrow P'$
        \State $M \leftarrow [] $
        \For{$(\theta_{i}, o_{i})\ in\ P$} 
            \State $\theta_{new}\leftarrow expand(\{\theta_{i}\})$ \Comment{expansion operator}
            \State $o_{new} \leftarrow evaluate(\theta_{new})$ \Comment{compute outcome}
            \State $M \leftarrow M + [(\theta_{new}, o_{new})]$
        \EndFor
        \State $A \leftarrow A + sample(M, N_{filter\ archive})$ \Comment{archive filtering} 
    \EndWhile
\end{algorithmic}
\end{algorithm}

\end{document}